\documentclass{article}
\usepackage{float}
\usepackage{iclr2027_conference}
\usepackage{times}
\usepackage[T1]{fontenc}
\usepackage{amsmath,amssymb,booktabs,multirow,graphicx}
\usepackage{hyperref,url,xcolor,microtype}
\DeclareMathSizes{10}{10}{7}{7}
\newcommand{\method}{\textsc{CoordinateClosure}}
\newcommand{\bench}{\textsc{LifecycleSkillBench}}
\hypersetup{hidelinks}
\title{Agent Skills Under Change: A Lifecycle Benchmark and Coordinate-Guided Updates}
\author{Anonymous Authors}
\AtBeginDocument{%
  \iclrfinalcopy
  \author{%
    Wei Yuan$^{1}$ \quad Qianghuai Jia$^{1}$ \quad Guanhua Chen$^{2}$ \quad Piji Li$^{3}$\\
    \textbf{Longyue Wang$^{1}$ \quad Zhao Xu$^{1}$ \quad Weihua Luo$^{1}$}\\[0.6ex]
    \normalfont\small $^{1}$Alibaba Group\\
    \normalfont\small $^{2}$Southern University of Science and Technology\\
    \normalfont\small $^{3}$Nanjing University of Aeronautics and Astronautics
  }
  \hypersetup{pdfauthor={Wei Yuan, Qianghuai Jia, Guanhua Chen, Piji Li, Longyue Wang, Zhao Xu, Weihua Luo},pdftitle={Agent Skills Under Change: A Lifecycle Benchmark and Coordinate-Guided Updates}}
  \let\arxivOriginalMaketitle\maketitle
  \renewcommand{\maketitle}{\arxivOriginalMaketitle\lhead{Preprint}}
}

\begin{document}
\maketitle

\begin{abstract}
\hyphenpenalty=3000
As agents accumulate reusable Skills, their capabilities depend on artifacts that must evolve with changing facts, tools, and policies. Yet an update that fixes today's failure can silently break working behavior and propagate errors into future tasks. We introduce \bench{}, a lifecycle evaluation protocol and development benchmark that pairs reproducible environmental changes with affected and protected tasks and tracks recovery, retention, and error propagation across updates. We also propose \method{}, a coordinate-guided update framework spanning four dimensions: facts, interfaces, execution boundaries, and policies. It localizes edits, checks dependencies, and combines public repair feedback with execution replay to decide whether to adopt an update. In controlled development experiments with matched public inputs, model-call schedules, and output caps, coordinate-structured patches improve first-pass state-check acceptance from 68.8\% to 98.4\% for Qwen3.7-Plus and from 76.6\% to 85.9\% for GLM-5.2. In a separate study of four 16-step maintenance streams, the complete controller, including repair and checkpoint restoration, preserves the expected state in 64/64 positions, compared with 56/64 for whole-state rewriting. Recorded input and output tokens fall from 958,833 to 42,225, a 95.6\% reduction. These experiments show how lifecycle evaluation exposes error persistence across updates, while coordinate-structured patches improve first-pass acceptance and the full CoordinateClosure controller reduces recorded token use relative to whole-state rewriting in the tested maintenance settings.
\end{abstract}

\section{Introduction}
\suppressfloats[t]
\begin{figure}[t]
\centering
\includegraphics[width=\linewidth,trim=24bp 19bp 19bp 20bp,clip]{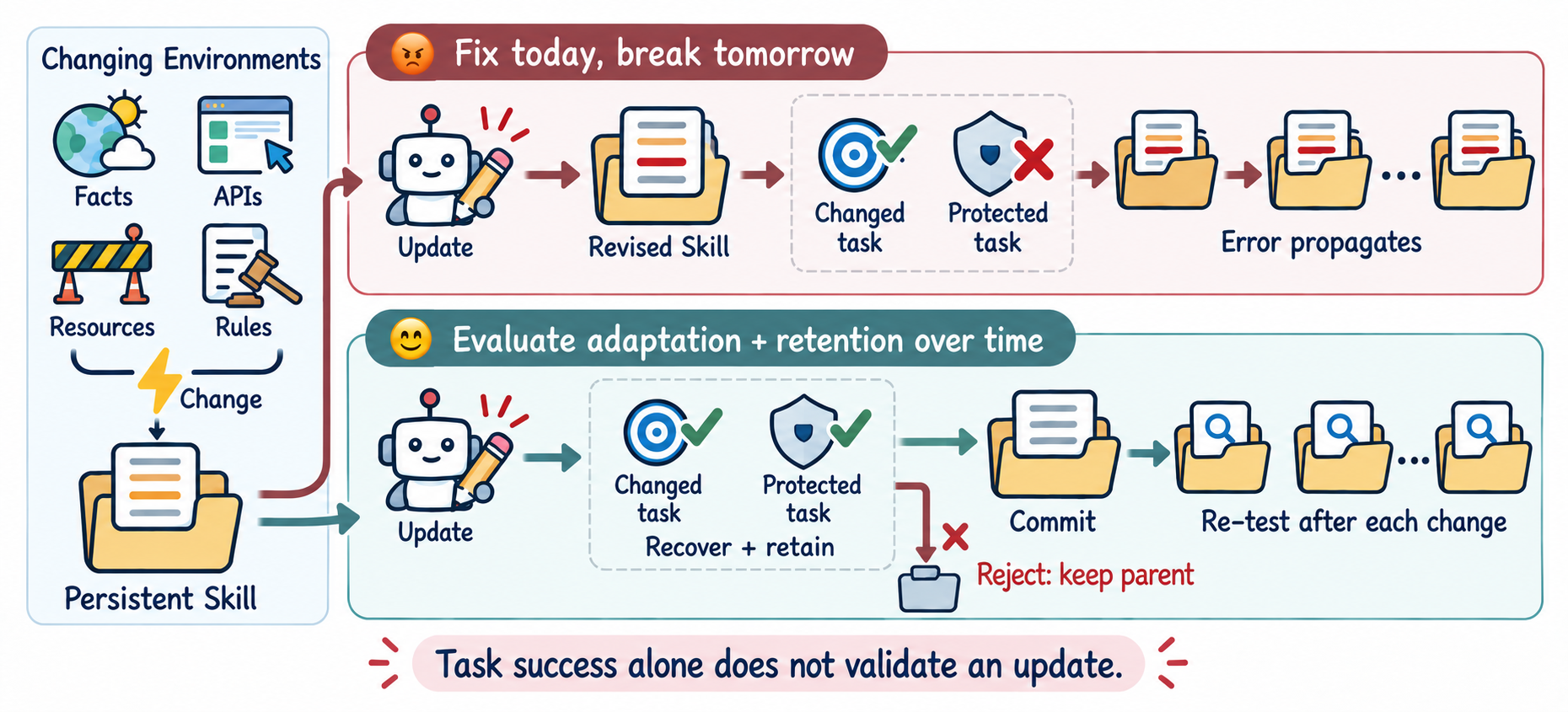}
\caption{Risks across successive Skill updates. An edit can repair the current task while breaking existing capabilities, with errors persisting into later tasks. Checking recovery and retention after each change informs whether to adopt the new version.}
\label{fig:lifecycle-motivation}
\end{figure}
Agent Skills store successful procedures, tool-use rules, and domain knowledge for reuse across tasks. Facts expire, APIs migrate, and execution requirements and policies change. Updating this shared state requires repairing affected behavior while preserving the capabilities that future tasks rely on.

Agents can accumulate reusable skills and maintain evolving strategies~\citep{wang2023voyager,zhang2026ace}, while lifelong benchmarks examine learning and skill evolution across sequential tasks~\citep{zheng2025lifelongagentbench,zhang2026skillflow}. We focus on updates triggered by environmental changes: can a revised Skill recover affected tasks while preserving capabilities that remain valid, and how do its effects carry into later versions? Answering this question requires linking each change to affected and protected tasks across update histories. We examine three update failures: edits to the wrong state region, inconsistent dependencies, and constraint compliance without goal completion.

\begin{samepage}
Our two main contributions are a lifecycle evaluation protocol with executable development cases and a coordinate-guided update method:
\begin{enumerate}
\item \bench{}, which pairs reproducible environmental changes with affected and protected tasks, tracking recovery, retention, error propagation, and execution cost across successive updates.
\item \method{}, which localizes Skill edits along four change dimensions, checks dependencies, and combines public repair feedback with execution replay to decide whether to adopt an update.
\end{enumerate}
\end{samepage}

Under matched public inputs, model-call schedules, output caps, and repair procedures, coordinate patches improve first-pass acceptance over generic patches by 29.7 percentage points for Qwen3.7-Plus and 9.4 for GLM-5.2, with similar generation token use. In a separate four-stream maintenance study, the complete controller, including scoped input selection, rollback, and deterministic repair, raises the fraction of states passing the expected-state checks from 87.5\% to 100\% relative to whole-state rewriting, with 95.6\% fewer recorded input and output tokens.

\section{Problem Definition}
Let $S_t$ denote the persistent Skill artifact before event $e_t$, and let $E_t$ contain the evidence available to the updater. The candidate is $\hat S_{t+1}=U(S_t,e_t,E_t)$. The evaluator holds affected tasks $T_t$, which require adaptation, and protected tasks $P_t$, whose requirements remain unchanged.

We call an update successful when affected tasks recover, protected capabilities are retained, the new behavioral constraints hold, and required output artifacts are produced. A failure-to-change occurs when one or more of these obligations remain unsatisfied despite an attempted update. Rejection preserves the previous state but does not count as successful adaptation when the affected task remains unsolved.

Let $R_t,K_t,B_t,X_t\in\{0,1\}$ denote affected-task recovery, protected-task retention, behavioral compliance, and executable delivery. $R_t$ and $K_t$ equal one only when every task in $T_t$ and $P_t$, respectively, passes its declared completion checks; $B_t$ and $X_t$ require all applicable behavioral obligations and required outputs across these tasks to pass. An unresolved check prevents its indicator from being one. Over $T$ evaluated updates, define
\begin{equation}
J_t=R_tK_tB_tX_t,\qquad
\mathrm{JointSafeCoverage}=\frac{1}{T}\sum_{t=1}^{T}J_t.
\end{equation}
Unknowns, timeouts, and parse failures contribute zero to this conservative success rate and remain separately labelled in the denominator. Component scores and exact-state agreement are reported separately; correlated stream states are not independent events.

\section{LifecycleSkillBench}
\begin{figure}[!htb]
\centering
\includegraphics[width=\linewidth,trim=99bp 26bp 50bp 39bp,clip]{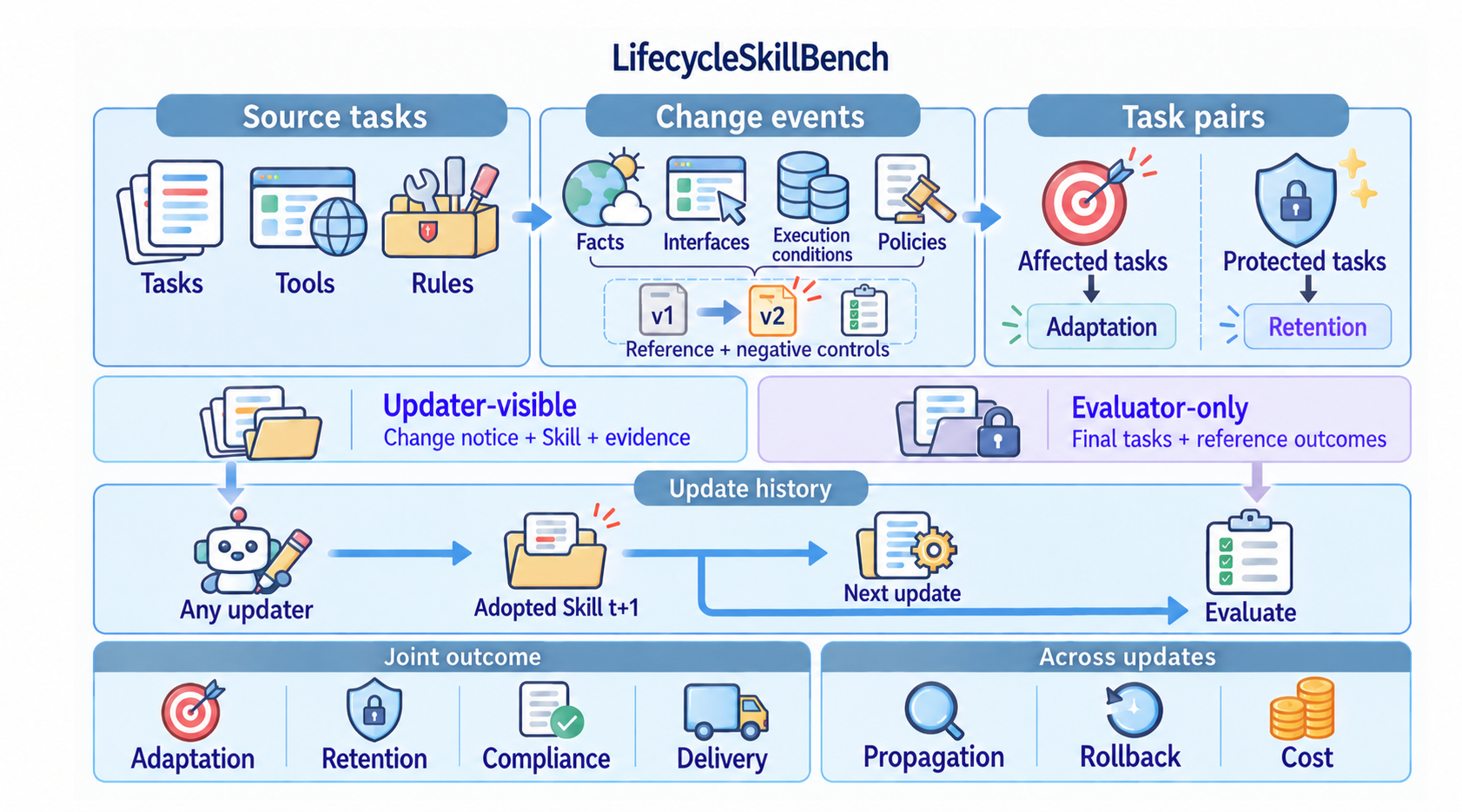}
\caption{LifecycleSkillBench construction and evaluation. Public tasks, tools, and rules support reproducible environmental changes, paired affected and capability-retention tasks, and persistent update histories. The evaluated method receives the change description, pre-update Skill, and available evidence. The evaluator separately stores test tasks, reference outcomes, and recoverable historical versions, and reports recovery, retention, error propagation, and execution cost.}
\label{fig:benchmark-protocol}
\end{figure}
LifecycleSkillBench evaluates whether reusable Skills remain effective as their operating environment changes (Figure~\ref{fig:benchmark-protocol}). Each event specifies the before/after environment and pairs affected tasks with tasks whose requirements remain unchanged, testing both recovery and regression. Successive events reuse the adopted Skill to track persistent errors. Appendix~\ref{app:design-rationale} explains the design and coverage audit.

\subsection{Evaluation objectives}
We ask whether disrupted tasks recover, unaffected tasks still succeed, and an edit's errors persist into later tasks. These define task recovery, capability retention, and error propagation. For example, adapting a broken tool call to changed parameters and completing the same user request constitutes recovery.

\subsection{Event construction and splits}
Each event records before/after environment configurations, a change description, and available update evidence. Affected tasks require adaptation; protected tasks retain valid goals and check existing capabilities. The evaluator stores reference outcomes and scoring rules. Execution records retain tool calls, final states, errors, and costs; reference outcomes and final-test feedback are withheld from proposal and repair.

The benchmark specifies source-grouped development, calibration, and final-evaluation splits, keeping related variants together. Development data support method design, calibration data determine check configurations, and final-evaluation data compare frozen methods. These are protocol requirements; the reported batches' actual partitions and uses are summarized in Appendix~\ref{app:batch-splits}. Tasks used for debugging remain development data.

\paragraph{Event validity.}
A valid event binds its source version, scoring and unknown-handling rules, and reconstruction procedure. Both a working reference and discriminative negative controls are required. If a changed policy makes the goal infeasible, compliant refusal is a stress-test outcome, not adaptation. Event validity is independent of method rankings.

\subsection{Data sources and event construction}
We organize 385 event-construction specifications across ToolSandbox (300), TimeWarp (36), AgentDojo (27), SkillsBench (16), and cross-source combinations (6). Figure~\ref{fig:coverage} summarizes construction, executable coverage, and protection checks; source rules appear in Appendix~\ref{app:inventory}, Table~\ref{tab:source-composition}.

\paragraph{Executable development coverage.}
The reliability bundle contains 19 development conditions (12 ToolSandbox, six historical IANA rule changes, and one synthetic retail-policy condition), with 201 control records including 26 unknown controls. A separate 12-step native-tool stream checks continuity. These are reference/control checks, not learned-method successes or independent test events (Appendix~\ref{app:reliability-update}).

\begin{figure}[!htb]
\centering
\includegraphics[width=\linewidth]{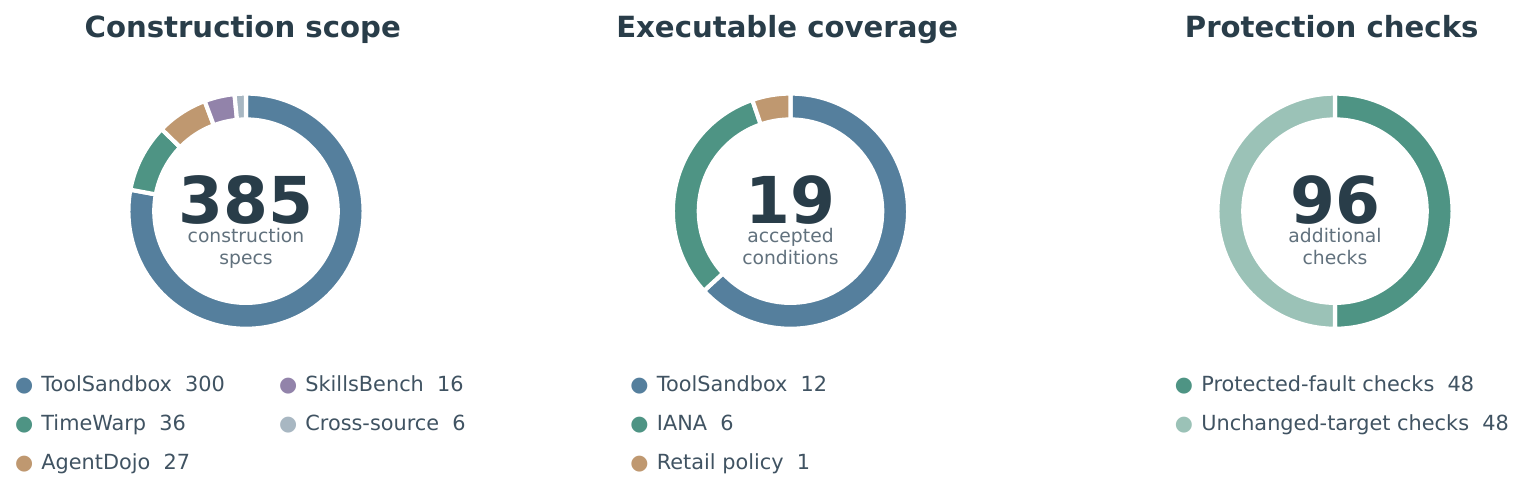}
\caption{Construction scope and validation depth. Specifications, accepted conditions, control records, and correlated update steps are distinct units, not a pooled sample count.}
\label{fig:coverage}
\end{figure}
Beyond the 201 baseline control records, 96 additional checks test protected-behavior discrimination: 48 deliberately faulty protected executions and 48 unchanged-target executions on the same artifacts. These checks deepen validation of existing conditions rather than add independent events. The separate stream checks error carryover and rollback; local protection checks do not establish full-dialogue retention.

\subsection{Successive updates}
Each adopted Skill becomes the next update's starting point, exposing accumulated errors. The protocol can also test unnecessary edits without environmental change (no-change), readaptation when an earlier rule returns (reversion), and reversal of a failed update (rollback). These name test conditions, not completed sample counts. Independent source events, trajectories, correlated update steps, and repeated trials are counted separately. Appendix~\ref{app:inventory} summarizes construction and source-grouping rules; the supplementary manifests bind reconstruction and scoring records.

\section{CoordinateClosure}
\subsection{Coordinates connect edit scope to verification}
\begin{figure}[!htb]
\centering
\includegraphics[width=\linewidth,trim=45bp 14bp 44bp 3bp,clip]{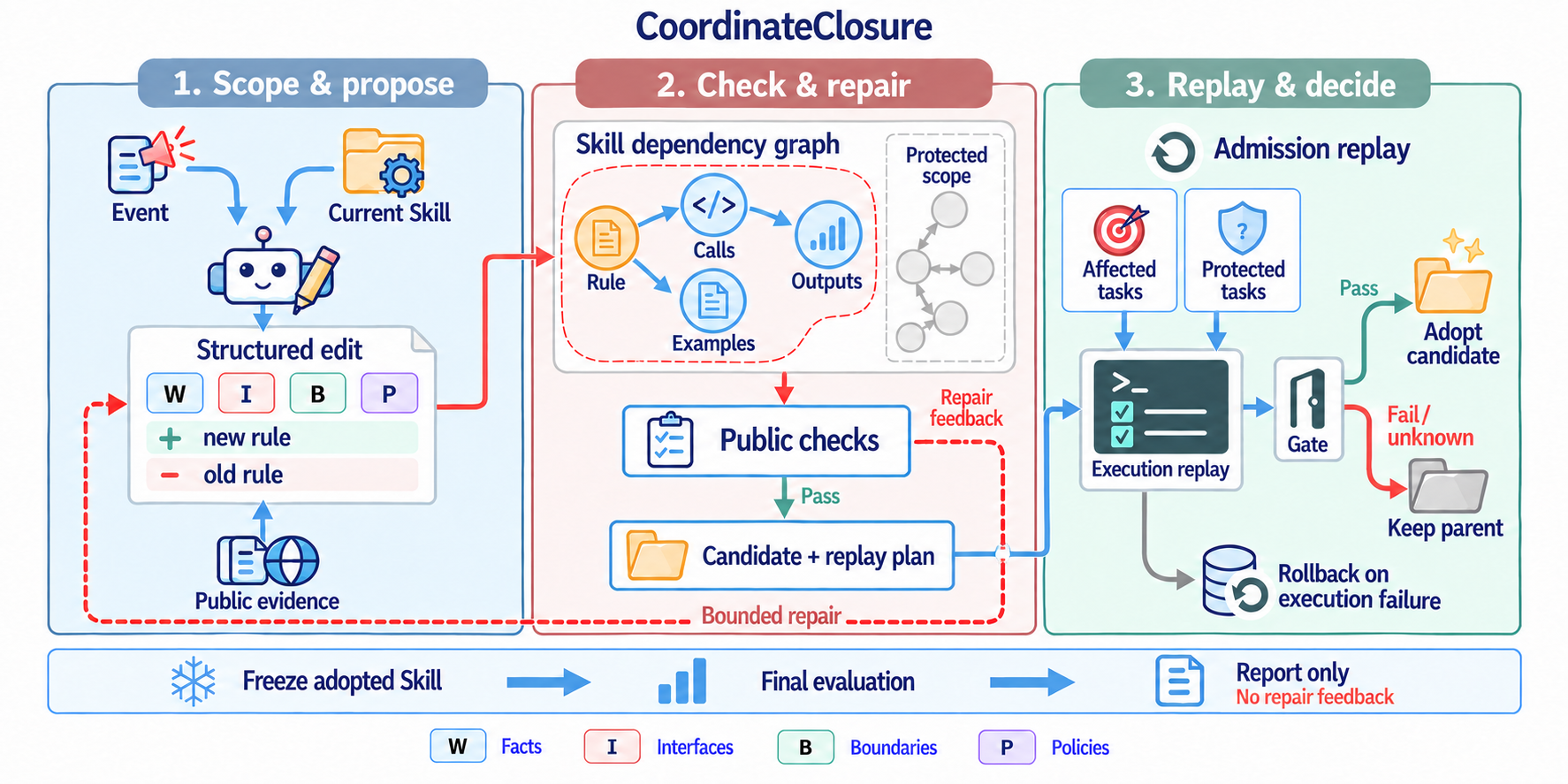}
\caption{CoordinateClosure update workflow. Changes guide edits and dependency checks, followed by affected-task and capability-retention replay. Public-check failures may trigger repair; failed or unknown execution checks retain the old version. The lower track evaluates frozen updates without repair feedback. W/I/B/P denote facts, interfaces, execution boundaries, and behavioral policies.}
\label{fig:method}
\end{figure}

We organize requirements into facts and provenance (world), tool signatures (interface), runtime and delivery contracts (boundary), and behavioral obligations (policy). An event can affect several domains. A structured (typed) patch records the edit location, replacement, and protected scope; domain labels link edits to checks, and dependency expansion identifies related requirements. Models propose and repair candidates; deterministic checks govern admission (Figure~\ref{fig:method}).

\subsection{From a change notice to an adopted update}
The updater receives the current Skill and a public change notice. It first identifies which facts, interfaces, execution conditions, or policies are affected and proposes edits to the corresponding locations. A coordinate names the kind of requirement; a path identifies its location in the Skill; the replacement specifies the proposed change. Protected locations define what must remain valid. This separates deciding where a change belongs from regenerating the whole artifact.

For example, suppose a payment API changes both its argument name and its amount unit. The interface coordinate identifies the affected call binding, while the update must also adjust the amount conversion and any examples using the old convention. If the return contract changes as well, result interpretation joins the checking scope. Unrelated payment rules remain protected. A single event can therefore induce several linked edits: coordinates organize their responsibilities, and declared dependencies connect their checks. The current checker uses explicit contract rules; it does not infer arbitrary semantic dependencies.

The controller then follows declared dependencies to check related rules, calls, and examples. A failed public check produces repair feedback containing the error location and violated requirement. Within the repair budget, the updater revises the candidate. Replay executes affected and protected tasks with the resulting Skill; adoption requires recovery, retention, compliance, and delivery to pass together. A failed or unknown check retains the parent. If execution requires rollback, the controller restores a verified checkpoint. Final evaluation tests the frozen adopted Skill without returning feedback to the updater. Appendix~\ref{app:design-rationale} describes version binding and decision records.

\paragraph{Dependency expansion.}
An edit can affect related content: changing a refund rule may require checking call arguments, examples, and output checks. Starting from directly edited items $C_j$, we follow declared dependencies in $D_j$. An edge $u\rightarrow v$ means that editing $u$ requires checking $v$. The resulting dependency closure is
\begin{equation}
\mathrm{Closure}_j(C_j)=C_j\cup\{v:\exists u\in C_j,\ u\leadsto v\text{ in }D_j\}.
\end{equation}
Items are rewritten only when needed. Closure terminates on the finite graph and is idempotent. Replay then tests affected and protected tasks and can expose execution errors missed by the declared dependencies.

\section{Experiments}
We first validate event execution and fault detection, then compare first-pass edits, continual-maintenance cost, and admission efficiency. Table~\ref{tab:controlled} summarizes the controlled studies; Appendix~\ref{app:core-audit} gives configurations and cost accounting.

\paragraph{Research questions.}
\textbf{RQ1 (Benchmark reliability):} Are events reproducible, and do checks detect intended failures and regressions?
\textbf{RQ2 (First-pass edits):} Do coordinate-structured patches improve initial acceptance under matched generation settings?
\textbf{RQ3 (Continual maintenance):} Does the complete controller reduce persistent errors and token use?
\textbf{RQ4 (Admission efficiency):} Can dependency filtering reduce replay executions while preserving admission decisions? Supplementary diagnostics examine repair and distinguish compliance from task completion (Appendix~\ref{app:diagnostics}).

\subsection{Experimental setup}
\paragraph{Units and comparisons.}
An event is an environmental transition; a task is a user goal; a trajectory records its dialogue and tool calls. The 385 construction specifications define the intended inventory; execution validation covers 19 development conditions. The method comparisons below use separate controlled fixtures. Controlled comparisons use four 16-step development streams per arm and backbone. Typed and generic patches share public inputs, a 2,048-token output cap, and repair procedures. Appendix~\ref{app:worked-case} gives model settings, scoring procedures, and a worked example; Appendix~\ref{app:diagnostics} reports policy diagnostics.

\paragraph{Implemented comparisons.}
We implement three update strategies in a common harness: whole-state rewriting generates a complete replacement; generic local patches emit changes without coordinate labels; coordinate patches add change-domain and scope constraints. Full CoordinateClosure additionally combines dependency checks, repair feedback, execution replay, and checkpoint restoration. Table~\ref{tab:controlled} groups these comparisons and the replay-only versus dependency-filtered admission ablation.

\begin{table}[!htb]
\caption{Completed controlled development comparisons. A and B each use four 16-state streams; C reports the identical pattern on one retail and two calculation fixtures, each with 17 constructed candidates. Different baselines and units are not pooled into an overall ranking.}
\label{tab:controlled}
\centering\small
\begin{tabular}{p{2.5in}rr}
\toprule
Metric & Baseline strategy & Evaluated variant \\
\midrule
\textit{A. First-pass updates} & Generic patch & Coordinate patch \\
Qwen initial state checks & 44/64 & \textbf{63/64} \\
GLM initial state checks & 49/64 & \textbf{55/64} \\
Qwen repair-required updates & 20 & \textbf{1} \\
GLM repair-required updates & 15 & \textbf{9} \\
Qwen final, same deterministic repair & 64/64 & 64/64 \\
GLM final, same deterministic repair & 64/64 & 64/64 \\
\midrule
\textit{B. Continual maintenance} & Whole-state rewrite & CoordinateClosure \\
\multicolumn{3}{l}{\footnotesize CoordinateClosure includes repair and checkpoint restoration.} \\
Stream exact-safe states & 56/64 & \textbf{64/64} \\
Observed stream total tokens & 958,833 & \textbf{42,225} \\
\midrule
\textit{C. Admission checks (per fixture)} & Replay only & \shortstack[r]{Dependency-filtered\\replay} \\
Admission workflow executions & 32 & \textbf{8} \\
Invalid candidates rejected & 15/15 & 15/15 \\
Joint successes & 1/17 & 1/17 \\
\bottomrule
\end{tabular}
\end{table}

\paragraph{Admission versus evaluation.}
Admission tasks decide whether to adopt an update; final tasks assess recovery and retention after freezing, without feeding back into repair. This is a formal protocol requirement, not a property established for all development runs. Database state, required actions, compliance, and goal completion are checked separately. All planned runs remain in the denominator. A joint outcome is unknown only when no required check has failed and at least one remains unresolved.

\paragraph{Statistical reporting.}
Specifications, candidates, trajectories, and scored outcomes are counted separately. With $y$ successes and $u$ unknowns among $n$ runs, $[y/n,(y+u)/n]$ gives bounds under alternative resolutions of unknown outcomes. These bounds do not quantify sampling uncertainty. Source-clustered bootstrap intervals are reported for the paired policy-state diagnostic in Appendix~\ref{app:diagnostics}; the main controlled comparisons are descriptive results over four correlated streams per arm. They do not establish significance across independent tasks or robustness across model seeds. Table~\ref{tab:controlled} reports state-check acceptance and exact-state preservation; these scores do not instantiate full-task JointSafeCoverage.

\subsection{RQ1: Benchmark execution and fault detection}
Separate isolated installations reproduce 216 ToolSandbox controls and nine synthetic-retail outcomes, including two expected unknowns (Appendix~\ref{app:latest-diagnostics}).

Each event binds source versions, before/after requirements, affected and protected tasks, and scoring rules. Working references test feasibility, obsolete or deliberately incorrect solutions test fault detection, and injected operational failures test unknown handling. The 19-condition bundle contains 201 control records, including 26 unknown controls; a separate 12-step native-tool stream tests state carryover (Appendix~\ref{app:reliability-update}).

Across the 12 ToolSandbox conditions, the scorer detects 48/48 injected protected-branch errors, and all 48 paired unchanged-target controls pass. Together, reconstruction and these 96 additional controls support benchmark reliability on the covered conditions: execution outcomes are reproducible, and scoring distinguishes injected protected-behavior faults from unchanged behavior.

\subsection{RQ2: First-pass acceptance and repair demand}
\begin{figure}[!htb]
\centering
\includegraphics[width=\linewidth]{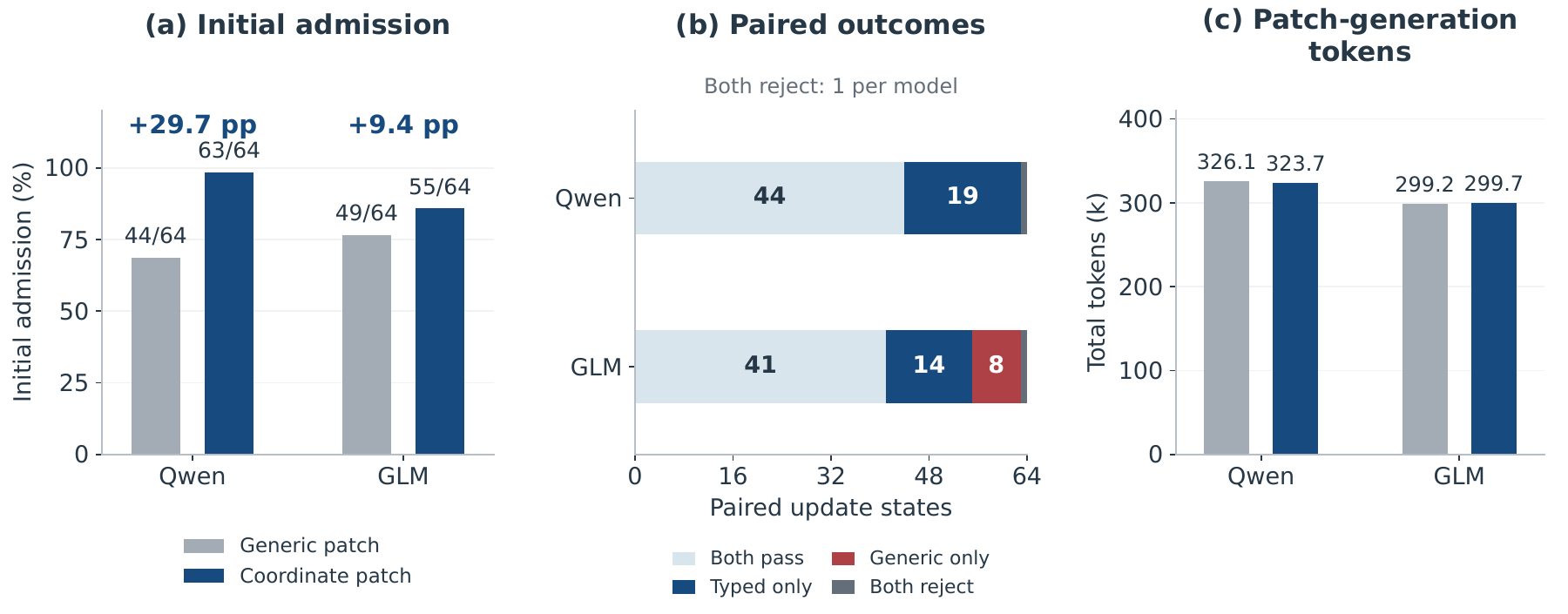}
\caption{Coordinate patches pass more first checks at similar recorded generation cost. (a) Acceptance before repair. (b) Blue: only coordinate patches pass; red: only generic patches pass. GLM gains 14 states and loses eight, a net gain of six. (c) Generation tokens are similar. Each arm contains four 16-state development streams (52 model-generated updates and 12 rollbacks); shared repair brings both formats to 64/64.}
\label{fig:first-pass}
\end{figure}
Coordinate patches require fewer repairs while using similar numbers of generation tokens: repair-required states fall from 20 to one for Qwen and from 15 to nine for GLM (Figure~\ref{fig:first-pass}). Evidence, call schedules, 2,048-token output caps, retries, checks, and repair procedures are matched. Coordinate patches specify the edit's dimension, original content, and replacement; generic patches lack coordinate-type constraints. The comparison therefore measures representation together with its parser and scope constraints.

Before repair, acceptance rises from 44/64 to 63/64 for Qwen and from 49/64 to 55/64 for GLM. Qwen has 19 coordinate-only passes and no generic-only passes; GLM has 14 and eight, respectively. Thus GLM improves in aggregate, although generic patches pass some states that coordinate patches fail. Each arm contains four correlated 16-state development streams: 52 model-generated updates and 12 deterministic rollbacks. On generated updates alone, counts are 32/52 versus 51/52 for Qwen and 37/52 versus 43/52 for GLM. Shared deterministic repair brings both formats to 64/64; the measured benefit is fewer states requiring repair, with tied final acceptance.

Across 52 generated updates per arm, parse failures fall from 11 to one for Qwen and from ten to four for GLM. This pattern is consistent with improved format adherence; acceptance also depends on state and scope checks. The experiment does not separate coordinate semantics from serialization or familiarity with the requested format.

\subsection{RQ3: Error persistence and maintenance cost}
In these four streams, the complete controller avoids the persistent state error observed with whole-state rewriting and uses fewer recorded tokens.

\begin{figure}[H]
\centering
\includegraphics[width=\linewidth]{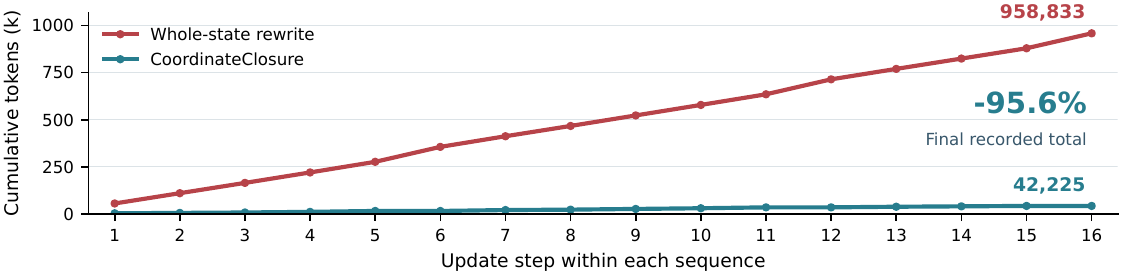}
\caption{Full controller vs.\ whole-state rewriting: cumulative recorded input and output tokens across four 16-step Qwen streams. The controller uses 42,225 tokens versus 958,833 (95.6\% less). State-by-state outcomes appear in Appendix Figure~\ref{fig:lifecycle-states}.}
\label{fig:lifecycle-cost}
\end{figure}
In a separate Qwen development study with four 16-step streams, the coordinate controller reaches 64/64 exact-safe states, compared with 56/64 for whole-state rewriting. One whole-state generation error persists through steps 8--15 and contaminates a rollback checkpoint. The eight failing states therefore represent one propagated error, not eight independent failures. The benchmark thus detects damage that would be missed by checking only the latest edit's target. The controller passes 63/64 initial state checks, including 12 deterministic rollbacks, and repairs one rejected proposal using deterministic rules derived from the public requirements. Retention here measures exact preservation of unchanged state fields; retention on complete tasks is evaluated separately.

The controller records 42,225 input and output tokens, compared with 958,833 for whole-state rewriting (95.6\% less). Rewriting reads and emits the full state and calls the model on all 64 steps; the controller supplies only the current event, generates scoped edits on 52 steps, restores checkpoints directly on 12 rollbacks, and applies public checks and deterministic repair. This comparison evaluates the combined input scope, representation, rollback schedule, and repair procedure; it does not isolate the benefit of coordinates. One additional failed rewrite request has no usage record; totals therefore represent observed usage. Appendix~\ref{app:core-audit} details accounting and configurations.

These savings arise in large-state, small-change maintenance. For example, an interface event changes \texttt{years} from integer to string, requiring \texttt{20} to become \texttt{"20"}. A scoped patch updates the call without regenerating unrelated tool definitions, templates, and facts. Together with direct checkpoint restoration, this reduces repeated generation across updates.

Task-level audits complement these state checks: one protected-task dialogue reports incorrect savings despite correct database actions (Appendix~\ref{app:retention-development}).

\subsection{RQ4: Admission efficiency}
\label{sec:mechanism-ablation}
Dependency filtering reduces admission replays while preserving decisions within the tested finite contracts. We compare four checking configurations on the same 17 constructed candidates for a retail interface migration: all 16 combinations of endpoint version, two argument-name versions, and protected-operation behavior, plus one missing candidate. The dependency checker tests the public endpoint--argument contract before execution. Rules and candidate hashes are frozen before this run; two admission workflows are separate from two final workflows. This is a development component ablation.

The dependency checker detects all 12 endpoint--argument inconsistencies. Adding it before replay preserves every admission decision while reducing admission workflow executions from 32 to 8 (75\%). Replay rejects all 15 candidates that fail final joint requirements; the one jointly valid candidate is accepted. Static checks therefore remove known inconsistencies cheaply, while replay checks task behavior beyond the declared dependency contract. All four configurations have joint success of 1/17: filtering protects against invalid updates but does not repair them.

Two additional ToolSandbox-derived calculation fixtures reproduce the same component pattern: dependency filtering reduces admission workflow slots from 32 to 8 per fixture while every arm remains at 1/17 joint success. Appendix~\ref{app:mechanism-ablation} reports all outcomes and execution-cost accounting.

\section{Related Work}
\textbf{Skill acquisition and evolution.} Voyager builds a reusable skill library, SkillWeaver discovers and refines skills as APIs, and ACE incrementally maintains strategy playbooks~\citep{wang2023voyager,zheng2025skillweaver,zhang2026ace}. Reflexion and Self-Refine use feedback to improve behavior or outputs~\citep{shinn2023reflexion,madaan2023selfrefine}. CoordinateClosure connects scoped edits to dependency checks and affected/protected-task replay, making version adoption an explicit decision.

\textbf{Lifelong agent evaluation.} LifelongAgentBench evaluates learning across interdependent interactive tasks; SkillFlow studies sequential skill discovery, patching, and reuse~\citep{zheng2025lifelongagentbench,zhang2026skillflow}. LifecycleSkillBench focuses on environmental transitions, pairing affected and protected tasks to track recovery, retention, and error propagation across adopted versions. This adaptation--retention objective also connects to continual learning and model editing; our update targets are external Skill artifacts (Appendix~\ref{app:additional-related}).

\textbf{Execution-based evaluation and update validation.} BFCL, $\tau^2$-Bench, ToolSandbox, and AgentDojo evaluate tool calls, user goals, stateful interaction, and adversarial behavior~\citep{patil2025bfcl,barres2025tau2,lu2024toolsandbox,debenedetti2024agentdojo}. These environments support execution-grounded assessment. Our protocol links selected sources to reproducible changes, paired task checks, and persistent version histories; the controller uses admission replay to decide whether to adopt a candidate.

\section{Discussion and Future Work}
The controlled studies show three complementary benefits: coordinate patches improve first-pass acceptance at similar generation cost, the complete controller reduces recorded token use relative to whole-state rewriting, and dependency filtering reduces admission replays without changing decisions. The observed maintenance savings apply to large Skill states undergoing small changes: scoped edits and checkpoint restoration avoid regenerating unchanged content. Together, these results show how structured updates and execution checks can reduce the work required to maintain reusable Skills across changes.

These findings concern small, partly development-visible settings. The public transition language fully specifies the expected state, so deterministic repair can recover every state without learned inference. Native interface pilots also show that underspecified format contracts can create apparent method differences that disappear after uniform normalization (Appendix~\ref{app:latest-diagnostics}). The main patch-format study uses two backbones; the maintenance comparison uses one. Source-disjoint evaluation with frozen contracts and additional backbones is needed to assess generalization beyond these fixtures. Future work should also measure full-dialogue capability retention, sampling uncertainty, and end-to-end cost and latency with complete usage records.

\begin{samepage}
\section{Conclusion}
Reusable Skills must evolve with their environments while preserving behavior that remains valid. This work contributes both an evaluation protocol for that process and a coordinate-guided update framework. LifecycleSkillBench pairs reproducible changes with affected and protected tasks and preserves update histories, making recovery, regression, and error propagation observable across versions. CoordinateClosure connects scoped edits and dependency checks to public repair feedback and execution-based adoption decisions.

The controlled studies reveal complementary benefits. Coordinate patches improve first-pass acceptance at similar generation cost, reducing repair demand. Across four maintenance streams, the complete controller preserves the expected state in 64/64 positions versus 56/64 for whole-state rewriting, with 95.6\% fewer recorded tokens. Dependency filtering reduces admission replays by 75\% while preserving decisions. Together, these findings show how evaluating successive updates and matching edit scope to change requirements can make Skill maintenance more reliable and economical in the tested settings.

\end{samepage}
\section*{Reproducibility Statement}
The ancillary file \texttt{supplementary.zip} provides core update code, frozen records for the main controlled comparisons, and an offline verification script. The script checks file integrity, recorded outcomes, and token accounting, and exercises the core update operations. Appendix~\ref{app:core-audit} details the main-result accounting. Separate native reconstruction experiments are reported in Appendix~\ref{app:latest-diagnostics}.

\section*{Ethics Statement}
All lifecycle events use public sources or simulated environments. No new human-subject data are collected.
Policy and transaction tests run in sandboxes and should not be interpreted as production safety certification.

\section*{AI Use Statement}
Generative AI assisted with literature research, experimental design and implementation, result analysis, manuscript drafting and revision, translation, and illustrations. Experimental model use is described in the relevant sections. The authors take responsibility for all content and evidence.

\bibliography{references,references_expanded}
\bibliographystyle{iclr2027_conference}
\appendix
\section{Updated benchmark reliability and artifact-level comparison}
\label{app:reliability-update}
\paragraph{Pinned factual-rule sources.}
The six core IANA conditions execute archived releases 2022a and 2023c; the four additional groups compare 2023c with 2025a~\citep{iana2022a,iana2023c,iana2025a}. Reconstruction uses the archive hashes in each frozen manifest, not the latest database. Release labels attached to individual changes identify when a rule or historical correction was recorded; they need not be the two archives executed in a comparison. In particular, the Vostok event tests a correction to the database's historical representation, not a claim that the old database offset describes the actual historical transition.

Supplementary controls retain 12 existing conditions: all 48 injected protected-branch errors are detected and all 48 paired unchanged-target checks pass. A further 120/120 historical positive/negative controls are reproduced. Only the first 96 checks are new; no conditions or model calls are added.

The development reliability bundle contains 19 conditions across 14 primary provenance groups: 12 ToolSandbox conditions from eight groups, six historical IANA rule changes from five jurisdiction groups, and one synthetic retail-policy condition. Shared protected tasks and variants preclude treating every condition as independent. Of 201 reference/control records, 26 deliberately test unknown handling: 12 ToolSandbox, 12 IANA, and two synthetic-retail controls. These counts exclude a separate 12-step tool-execution stream and all method-generation trials. The IANA workload executes historical time-zone rules rather than judging free-text factual answers. A separate Linux worker-isolation replay completes 60 conversions and 36 denied attempts to access evaluator material; this validates that runner, not the isolation of earlier experiments. These are development conditions, not a source-disjoint final test.

\paragraph{Two-model interface-update diagnostic.}
One further retail-tool migration uses separate admission and final tasks, two models (\texttt{qwen3.7-plus-2026-05-26}, \texttt{glm-5.2}), requested seeds 11 and 29, temperature zero, a 2,048-token output cap, and no repair. Twelve generation requests compare whole rewriting, untyped patches, and coordinate patches; the fourth arm reuses each coordinate candidate with admission replay. All candidates and decisions are frozen before final replay. The evaluator supplies reference workflows, so this tests generated call-binding artifacts rather than complete model-planned conversations.

For Qwen, affected-task passes are 2/2, 0/2, 2/2, and 2/2 across the four arms; all protected checks pass. The untyped proposals rename the endpoint but omit required argument renames. For GLM, the corresponding counts are 1/2, 2/2, 2/2, and 2/2; the remaining whole-rewrite candidate has an invalid schema, leaving its affected and protected outcomes unknown. Overall, 32 planned final slots contain 28 passes, two failures, and two unknowns, with 30 actual execution receipts. All eight admission checks pass. Independent raw-response parsing and native replay reproduce the records. Generation consumes 23,067 tokens; coordinate generation is slightly more expensive in this batch. This single-mechanism diagnostic supports an observed benefit of complete structured edits, not broad superiority, full-task retention, or an incremental gate advantage. Operationally failed earlier attempts remain archived separately.

\paragraph{Cross-source rule-version execution check.}
On the six accepted IANA development conditions, 36 generation requests compare the same three formats with two models and seed 11; the gated arm reuses typed candidates. Admission and final date/zone inputs are distinct. This experiment generates version-selection artifacts, not new time-zone rule implementations. All 24 admission conversions pass. Of 144 final slots, 126 execute and pass and 18 remain unknown: all six GLM whole-rewrite responses select the correct rule versions but also emit metadata rejected by the fixed schema. This is format nonconformance, not evidence of inferior factual adaptation. All valid methods tie, and the gate accepts every typed candidate. Independent recompilation of the official rules and replay reproduce all records. Total generation usage is 10,647 tokens. The result checks cross-source executability, not independent-test superiority or full conversational retention.

\paragraph{Separating format compliance from execution correctness.}
To test whether the unknown outcomes arise solely from extra descriptive fields, we conduct a post-hoc format-compatibility analysis. The same deterministic rule is applied to all 36 original responses: remove only fields outside the declared schema, without filling missing fields, changing required values, repairing invalid JSON, or consulting reference answers to modify candidates. The normalization policy and original-response hashes are fixed before rescoring. Six GLM whole-rewrite responses become executable; all 36 candidates are valid, and all 24 admission and 144 final conversions are actually executed and pass. The original strict result of 126 passes and 18 unknowns is retained separately. No additional model requests are made, so generation usage remains 10,647 tokens. This analysis shows that all methods select the correct rule versions on this workload and that the original difference concerns format compatibility, not factual adaptation. Because the rule was developed after inspecting the original results, this is reported as a sensitivity analysis rather than a replacement for the frozen scoring protocol.

\section{Design rationale and evidence for the two contributions}
\label{app:design-rationale}
We evaluate an update through its execution outcomes: it should recover affected tasks, satisfy the new behavioral constraints, and retain unaffected capabilities. Joint update checks assess whether these requirements hold together. To describe the scope of a change, we represent an event as
\begin{equation}
\Delta_t=(\Delta_t^w,\Delta_t^i,\Delta_t^b,\Delta_t^p),
\end{equation}
where the superscripts denote world facts, tool interfaces, execution boundaries, and behavioral policies. The world coordinate covers changing facts and their provenance, such as dates, entity attributes, and external knowledge; it excludes tool signatures, runtime resources, and behavioral rules. Coordinates localize edits and verification requirements; an event can activate several coordinates.

For example, changing a policy from five passengers per order to one makes a three-passenger booking invalid. If separate orders are permitted and authorized, the updated Skill must complete three valid bookings to satisfy the goal.
Operational errors, parse failures, timeouts, and unknown states remain in the denominator; unknown
is never silently counted as safe.

\paragraph{Why evaluate updates?}
A reusable Skill carries decisions from one task to the next. A task-level success can therefore hide two different costs: damage to another valid task and an error inherited by later versions. This motivates three connected design choices. Before/after environments establish what adaptation is required; affected/protected task pairs expose the trade-off within an update; persistent histories expose consequences across updates. These choices define a method-independent evaluation problem. Any updater can be evaluated through the same task and history interface.

\paragraph{Validity before scale.}
A valid event specifies the old environment, intervention, and new requirements; provides a working reference and appropriate negative controls; includes protected tasks; and binds scoring, unknown handling, provenance, versions, split, and reconstruction. Negative controls include stale solutions, no-ops, and damaged protected states. Infeasible goals are labelled policy stress tests. Together, these checks establish feasibility and scorer discrimination.

\paragraph{Coverage is an auditable profile.}
Coverage is recorded by change mechanism, source group, task obligation, and temporal condition, with evidence labelled specification-only, contract-level, or tool-executed. The earlier ToolSandbox-plus-policy audit covered 13 conditions; the expanded 19-condition accounting appears in Appendix~\ref{app:reliability-update}. Related variants share provenance, and the 385-specification inventory remains separate from executable validation.

\paragraph{Coordinates organize a larger update process.}
Coordinates label the responsibilities of edits and checks. Structured patches make the changed locations explicit; dependency expansion identifies related items to inspect; public checks identify concrete repair obligations; replay supplies the adoption evidence. The implemented workflow checker classifies failures including unavailable endpoints, argument names and types, unit conversion, resource prerequisites, result selection, export type, and edits to protected scope. This is a diagnostic classification, not an empirically validated learned classifier. Its dependency graph covers a declared finite contract language, not all semantic dependencies.

\paragraph{Repair information and decision records.}
The implemented repair packet contains the public contract, current candidate, raw proposal, errors with paths and explanations, dependency obligations, and a bounded repair instruction. Its purpose is to make the next edit answer a specific check failure. It excludes final reference answers and final-test feedback. Candidate/state hashes and evaluator versions link proposal, execution, and acceptance records. These records support decision tracing. The reported workflow study permits at most one learned repair with the same public feedback for each proposal format. Utility checks cover a narrower schema contract and should not be described as the full dependency compiler.

\paragraph{Version binding and proposal roles.}
Candidate bytes, the parent-state hash, the public evidence packet, and compiler/evaluator versions are bound by a deterministic identifier. Unknown operators, unsupported fields, parent drift, or expected-hash mismatch fail closed. Rejection leaves the parent unchanged; rollback returns to a hash-verified checkpoint. Where multiple agents propose or review typed claims, agreement cannot replace execution checks, and no new patch may be synthesized after admission without revalidation. This role separation is an implementation contract; the reported comparisons do not isolate a multi-agent benefit.

\subsection{From design choices to measurable benefits}
The design follows the failure path of an update: an incorrect edit scope, an omitted dependency, an unsuccessful repair, or an incorrect adoption decision. Table~\ref{tab:mechanism-evidence} makes each component's output and evaluation target explicit. These components serve complementary roles within the update procedure.
\begin{table}[!htb]
\centering\small
\caption{Implemented mechanisms and the evidence needed to assess them.}
\label{tab:mechanism-evidence}
\begin{tabular}{p{0.19\linewidth}p{0.34\linewidth}p{0.37\linewidth}}
\toprule
Mechanism & Operational output & Evaluation target and current evidence \\
\midrule
Responsibility labels and structured edits & Edit locations, owners, and replacements & First-pass contract acceptance; supported by the matched representation-and-validation comparison. \\
Dependency checks & Related requirements and identified contract violations & Missed obligations and protected-scope changes; evaluated within the declared finite contracts. \\
Diagnostic repair packet & Candidate, public contract, error locations, and repair requirements & Repair success and additional calls; shared-feedback workflow diagnostics, no exclusive method advantage established. \\
Replay admission & Adopt or retain the previous version, with execution receipts & False admissions and unnecessary rejections within the reported development fixtures. \\
Version-bound records & Candidate/state hashes and evaluator version & Decision traceability and reproducibility; these properties do not imply higher task success. \\
\bottomrule
\end{tabular}
\end{table}

\section{Historical development denominators and cost accounting}
\label{app:core-audit}
\begin{figure}[!htb]
\centering
\includegraphics[width=\linewidth]{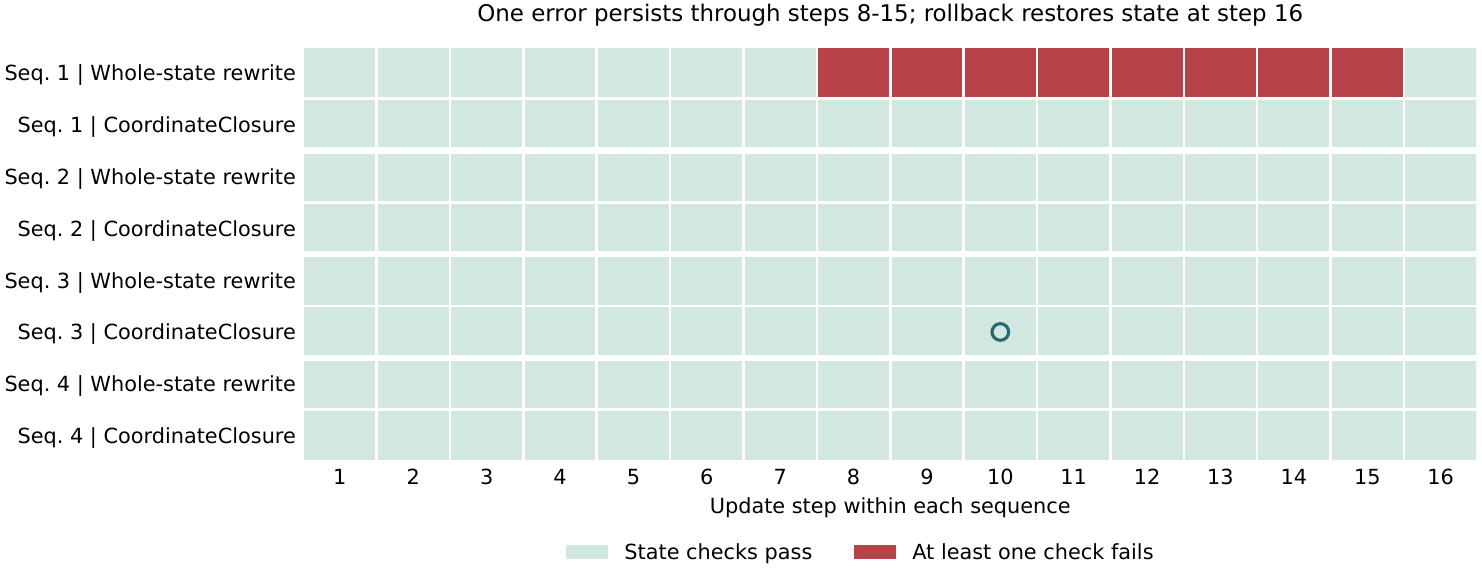}
\caption{State checks across four 16-step Qwen streams. One rewrite error persists through steps 8--15 of sequence 1; step 16 restores the initial state. The open circle marks deterministic repair at sequence 3, step 10. CoordinateClosure passes 64/64 states after repair, compared with 56/64 for rewriting; the eight failures share one error.}
\label{fig:lifecycle-states}
\end{figure}
\paragraph{Why local maintenance saves tokens.}
The state contains reusable tool definitions, call templates, and factual records, while an individual event changes only a small part. For example, one interface event changes the type of \texttt{years} from integer to string: the corresponding call must change from \texttt{20} to \texttt{"20"}, without regenerating unrelated entries. Scoped edits avoid repeatedly reading and emitting the full state; checkpoint restoration similarly avoids model-generated rollback. Recorded input tokens decrease from 517,999 to 30,414, and output tokens from 440,834 to 11,811. This identifies large-state, small-change maintenance as the setting in which the complete pipeline is particularly economical. It does not imply the same saving over another local-patch method; the matched representation comparison has similar token use.

Both studies use four 16-step development streams. States share histories and cannot be treated as 64 independent samples for significance tests. The eight incorrect rewrite states reflect one erroneous field persisting through steps 8--15. Controlled scores check changed fields, unchanged fields, and complete expected-state equality, rather than conversational success or general behavioral safety.

The matched-format study uses \texttt{qwen3.7-plus-2026-05-26} and \texttt{glm-5.2}, temperature 0, a 2,048-token output cap, 240-second timeout, at most two operational retries, and concurrency 8; no provider seed is controlled. Visible-input hashes match for all 64 paired states, with 52 successful model responses per arm. Generic/typed parse failures are 11/1 for Qwen and 10/4 for GLM, so initial acceptance includes serialization and constraint-check effects. Recorded total tokens are 326,059/323,747 for Qwen and 299,202/299,748 for GLM. Each generic arm has one additional failed request without usage; missing usage is not interpreted as zero cost.

The rewrite comparison uses the same Qwen version and an 8,192-token output cap, but differs in input scope, output representation, rollback scheduling, and repair. Rewrite input/output tokens are 517,999/440,834 versus 30,414/11,811 for the controller, with 64 versus 52 successful calls. Summed successful-request durations are 7,576.884 versus 248.370 seconds, not concurrent wall-clock latency; CPU checking, materialization, and repair are not separately timed.

The public event language fully specifies state transitions: a deterministic public interpreter reproduces all 64 expected states without a model. Perfect repaired results validate execution reliability within that language; first-pass contract adherence, repair demand, and state-copying cost remain separately observable. This does not show that a model can infer semantic obligations absent from the public evidence.

\section{Construction inventory and release checks}
\label{app:inventory}
Five manifests contain 385 specifications: 300 ToolSandbox, 36 TimeWarp, 27 AgentDojo, 16 SkillsBench, and six cross-source combinations. A specification describes construction and evaluation, not a completed model execution. Counts follow source-selection and variant-generation rules, not balanced sampling: ToolSandbox contributes 75 scenarios with four interface variants each. Related variants are not independent samples. Table~\ref{tab:source-composition} summarizes the construction mechanisms.

This inventory is reported separately from controlled studies, $\tau^2$ policy interactions, and SkillFlow execution diagnostics; their counts cannot simply be added. The experiments identify executed tasks, models, and trajectories per batch.

\begin{table}[!htb]
\centering\small
\caption{Source composition of the 385 construction specifications, grouped by the selected construction mechanisms.}
\label{tab:source-composition}
\begin{tabular}{p{1.0in}p{3.5in}r}
\toprule
Source & Construction mechanism & Specs. \\
\midrule
ToolSandbox & 75 tool-use scenarios, each with four interface variants & 300 \\
TimeWarp & Tasks paired across environment versions; factual changes require separate verification & 36 \\
AgentDojo & Versioned injection-task additions or updates & 27 \\
SkillsBench & Four source tasks, each with four execution-condition changes & 16 \\
Cross-source & Combinations reusing components from existing sources & 6 \\
\bottomrule
\end{tabular}
\end{table}
The five manifests use the following construction rules:
\begin{itemize}
\item ToolSandbox selects 75 source scenarios and creates four perturbations per scenario: tool description, argument type, argument description, and tool name. The 300 variants stay grouped in three splits of 25 source scenarios (100 variants) each.
\item TimeWarp binds 36 source tasks to environment versions 1 and 5. The manifest holds intent and backend content constant; a version change alone therefore does not demonstrate changed facts. Classification requires inspecting the actual change.
\item AgentDojo contributes 27 versioned injection-task updates or additions from 23 source groups. Not every record has a paired predecessor, and injection-task changes cannot all be interpreted as business-policy changes.
\item SkillsBench combines four source tasks with four execution-condition changes each, including resources, output contracts, and dependency versions, yielding 16 specifications.
\item Six cross-source specifications cover three pairwise combinations. Their components reuse existing sources rather than adding six independent source tasks.
\end{itemize}
Each experiment reports the executed subset of this inventory.

\paragraph{Ancillary code and records.}
The package contains core update code, frozen records for the main controlled comparisons, and offline verification tools. It supports recomputation of the packaged results and testing of the core update operations. Historical archive checks and separate native reconstruction experiments are reported in Appendix~\ref{app:latest-diagnostics}.
\paragraph{Earlier scoped reconstruction checkpoints.}
A 26-file workflow module supplies contracts, mutation recipes, a source-family-grouped development manifest, separate evaluation fixtures, a dependency checker, and a model-free execution entry point for four tool families and eight conditions. It pins the ToolSandbox revision and source hashes while excluding third-party raw data. A fresh macOS arm64 installation with Python 3.11.13 passes all 40 native reference checks and dependency checks.
A 37-file extension adds two numerical-tool fixtures, a provider-free adapter, reconstruction instructions, and a license ledger. Extraction verifies all hashes and 60 reference checks (40 workflow and 20 numerical). The archive excludes credentials, provider runners, and third-party raw data. 
\paragraph{Earlier source-module acceptance and exclusions.}
The scoped ToolSandbox module accepts 12 development conditions from eight source groups and five modules. Independent replay reproduces 120 native reference checks and 12 unknown-handling injections. Other candidates remain outside native acceptance: 18 of 20 BFCL cases pass call-contract checks without executing tools, while 24 StreamingQA cases support annotation reconstruction but lack verified factual transitions.

\subsection{Partitions and uses of the reported batches}\label{app:batch-splits}
The table distinguishes the benchmark's split protocol from the actual uses of the reported batches. Admission/final-task separation alone does not establish source-disjoint final evaluation.
\begin{table}[!htb]
\centering\small
\begin{tabular}{p{0.31\linewidth}p{0.61\linewidth}}
\toprule
Batch & Actual partition and evidence use\\
\midrule
19-condition reliability checks & Development reference, fault, and unknown-handling controls; checks execution and scoring.\\
Matched patches and continual maintenance & Four 16-step development streams; public checks and repair are visible; descriptive comparisons of correlated states.\\
Finite interface-dependency ablations & 17 constructed candidates per fixture; two admission and two final workflows are separate; development component tests.\\
Retail and airline workflow pilots & Specified admission and final task/user groups are disjoint; development diagnostics, not source-disjoint full-controller evaluation.\\
Dialogue and retention diagnostics & Tasks, models, and judgment scope recorded per batch; post-hoc audits and original outcomes retained separately.\\
\bottomrule
\end{tabular}
\end{table}

\section{Additional related work}
\label{app:additional-related}
This section expands the positioning in the main text: continual learning and editing motivate adaptation and retention, skill improvement informs candidate generation, and agent benchmarks supply execution settings. Further connections concern memory, repair, and behavioral testing. The literature scope and the implemented experimental comparisons are reported separately.
\paragraph{Continual learning, editing, and temporal knowledge.}
Learning without Forgetting, EWC, GEM, and episodic replay study acquisition of new capabilities
while retaining old ones \citep{li2016lwf,kirkpatrick2017ewc,lopezpaz2017gem,chaudhry2019replay}.
MEND, ROME, MEMIT, and SERAC investigate targeted edits and their scope
\citep{mitchell2021mend,meng2022rome,meng2023memit,mitchell2022serac}.
TENT adapts models at test time \citep{wang2021tent}; retrieval-augmented generation and
StreamingQA address external knowledge and temporal adaptation \citep{lewis2020rag,liska2022streamingqa}.
Our setting shares their adaptation--retention objective, but updates externally stored Skill artifacts
and evaluates tool behavior, execution state, and commit decisions. Our contribution is a joint protocol
for evaluating and controlling updates to external Skill artifacts.

\paragraph{Skill acquisition and candidate improvement.}
ReAct interleaves reasoning and action, Reflexion uses verbal feedback, and Voyager accumulates
reusable skills \citep{yao2023react,shinn2023reflexion,wang2023voyager}. SkillFlow directly studies
lifelong skill discovery and evolution \citep{zhang2026skillflow}; therefore skill evolution is an
existing research problem. Toolformer, Gorilla, and ToolLLM develop tool-use capabilities
\citep{schick2023toolformer,patil2023gorilla,qin2023toolllm}. Self-Refine revises outputs using
self-feedback \citep{madaan2023selfrefine}; APE, OPRO, DSPy, and TextGrad optimize instructions
or compound language-model systems \citep{zhou2022ape,yang2023opro,khattab2023dspy,yuksekgonul2024textgrad}.
These works motivate stronger updater baselines. Our comparison asks whether candidate improvements
survive changed-task execution and protected-task replay under matched information and budgets;
citing these methods does not imply that all have already been implemented or evaluated here.

\paragraph{Agent and tool-use benchmarks.}
BFCL, $\tau$-bench, $\tau^2$-Bench, ToolSandbox, and AgentDojo provide tool-call, conversational,
stateful, or adversarial evaluation \citep{patil2025bfcl,yao2024taubench,barres2025tau2,lu2024toolsandbox,debenedetti2024agentdojo}.
AgentBench and GAIA broaden task coverage \citep{liu2023agentbench,mialon2023gaia}.
ALFWorld and WebShop supply interactive environments \citep{shridhar2020alfworld,yao2022webshop};
Mind2Web, WebArena, VisualWebArena, and WorkArena evaluate web interaction
\citep{deng2023mind2web,zhou2023webarena,koh2024visualwebarena,drouin2024workarena}.
OSWorld and SWE-bench emphasize computer and software execution
\citep{xie2024osworld,jimenez2023swebench}. These are related evaluation settings, not all sources
used in our experiments. We extend selected sources with paired updates, protected tasks, and
commit/rollback records; scores remain lifecycle-derived rather than official leaderboard results.

\paragraph{Persistent memory, repair, and risk evaluation.}
MemGPT manages hierarchical memory for long contexts and multi-session interaction
\citep{packer2023memgpt}; we additionally ask whether modified persistent artifacts still satisfy task requirements.
SWE-agent studies agent-computer interfaces, while Agentless separates software repair into
localization, repair, and patch validation \citep{yang2024sweagent,xia2024agentless}, offering related designs for scoped edits and execution checks.
ToolEmu emulates tool execution and evaluates risks with language models \citep{ruan2023toolemu}.
These settings also motivate distinguishing simulated judgments, call-contract checks, and actual execution when reporting evidence.

\paragraph{Multi-agent collaboration.}
AutoGen supports conversational coordination, MetaGPT structures role-based workflows, and
multiagent debate exchanges competing responses \citep{wu2023autogen,hong2023metagpt,du2023debate}.
These mechanisms can supply candidate diversity. CoordinateClosure separates proposal and repair
from executable admission: agreement among agents does not substitute for a successful replay.
Matched call budgets help distinguish update mechanisms from extra inference.

\paragraph{Behavioral testing and trustworthy reporting.}
CheckList motivates evaluating specified behaviors beyond aggregate accuracy
\citep{ribeiro2020checklist}. Datasheets and Model Cards motivate explicit data provenance and
intended-use boundaries \citep{gebru2018datasheets,mitchell2018modelcards}.
Recommendations on significance testing and uncertainty in few-run evaluation motivate paired
comparisons and uncertainty reporting \citep{dror2018testing,agarwal2021rliable}.
For lifecycle evaluation, independent source events should be distinguished from correlated steps,
and operational failures kept visible. A statistical non-inferiority claim requires a prespecified
margin; a nonsignificant difference alone does not establish preservation. We do not claim such a test here.
\section{Execution settings and a worked development case}
\label{app:worked-case}
\paragraph{Execution and provenance.}
The following settings apply to the saved $\tau^2$ retail exchange batch. The actor's requested model identifier is \texttt{qwen-plus}, with trial seed 11, temperature 0.2, at most 1,024 output tokens per call, and 40 steps per trajectory. API timeout is 90 seconds, configured API retries are zero, task timeout is 600 seconds, and concurrency is four. The simulator uses lifecycle-v5. Frozen records bind tasks, candidates, and executed policies by hash. Four source tasks across three generated policies and one stale control yield 16 trajectories. A trial seed does not guarantee deterministic hosted API responses, and a requested alias is not an immutable server-side weight version.

The frozen v1b configurations request \texttt{qwen3.7-plus-2026-05-26} and \texttt{glm-5.2}. Both use temperature zero, a 2,048-token output cap, a 240-second timeout, a retry setting of two, one-second backoff, and concurrency eight. Each non-rollback step makes one model call per arm; rollback steps make none. Provider seeds are uncontrolled, so these are not multi-seed stability results. This budget generates edits, unlike the native actor's 1,024-token task-execution budget.

\paragraph{Input and transition.}
Source task 0 requests two exchanges from a delivered order: a full-size keyboard with clicky switches, preferring RGB but accepting no backlight if that combination is unavailable, and a thermostat compatible with Google Home. The bulletin prohibits gift-card settlement of exchange price differences. Because this task uses a credit card, it is a protected task expected to remain valid, not a changed task specifically testing a newly forbidden action. The updater receives public policy evidence; the original user goal and reference actions are preserved. The actual patch replaces the following settlement clause while retaining the same-product restriction:
\begin{quote}\small
Before: The user must provide a payment method to pay or receive refund of the price difference. If the user provides a gift card, it must have enough balance to cover the price difference.

After: The user must provide a non-gift-card payment method to pay or receive refund of the price difference. Gift-card settlement is no longer permitted for exchanges.
\end{quote}
The candidate assigns this edit to \texttt{policy}. This case illustrates policy-edit localization.

\paragraph{Recorded execution.}
Indices below are zero-based saved-message indices. These are selected evidence excerpts, not a regenerated demonstration.
\begin{enumerate}
\item Messages 2--9 identify the account by name and ZIP, retrieve the order, and inspect product variants.
\item Message 10 identifies the permitted no-backlight fallback, presents the thermostat replacement, computes refunds of \$3.17 and \$13.46 (total \$16.63), and requests confirmation.
\item Message 11 confirms only these two exchanges and the \$16.63 refund to the original credit card.
\item Message 12 calls \texttt{exchange\_delivered\_order\_items}, mapping item IDs 1151293680 and 4983901480 to 7706410293 and 7747408585. The payment method is \texttt{credit\_card\_9513926}.
\item Messages 13--14 contain the returned order record and completion report. Database and necessary-action checks pass; no gift-card exchange call is observed.
\end{enumerate}

\paragraph{Scoring and limits.}
The database scorer checks final state; the action guard checks required writes and arguments; the violation monitor checks exchange payment parameters. A separate post-hoc audit gives the same method-blind trajectory to \texttt{qwen-plus} and \texttt{glm-5.2}, with temperature zero, an 8,192-token output limit, thinking disabled, and concurrency two. Packets contain the user scenario, updated policy, original natural-language assertions, and two common added requirements: fulfill applicable requests and obey policy. Both judges mark this case's requirements and simulator fidelity as passing, producing a joint diagnostic pass. This is not human-calibrated accuracy or an exhaustive factual verification of every assistant statement.

This worked case covers input, patching, execution, and scoring. It has no separate admission tasks or complete gated arm; independent final evaluation therefore requires a different batch with frozen admission decisions.
\section{Execution and measurement diagnostics}\label{app:diagnostics}
Public-task diagnostics motivate the joint endpoint. In the updated-policy arm of the $\tau^2$ development cohort, 28/44 observable trajectories contain no explicit violation, while only 10/44 complete the native task. Eighteen therefore avoid observed violations without recovering the task. In SkillFlow, both compared program pipelines pass 8/9 native cases; the unresolved case produces no required output files despite a zero return code. These diagnostics motivate separate behavioral and delivery checks.

\paragraph{Repair of rejected first-pass candidates.}
On 45 rejected development candidates frozen after the first-pass runs, obligation-grounded learned repair recovers 24/45 (53.3\%); deterministic repair using known public transition rules recovers 45/45. The latter is a rule-based upper bound. The selected candidates are correlated and are counted separately from independent events.

A subsequent end-to-end development run separates two admission tasks from two final tasks and records candidate, executed-policy, and gate hashes. All ten final trajectories contain the required exchange execution evidence; seven pass the joint diagnostic checks and three remain unresolved. The gate retains the parent policy. This establishes an executable audit path, while method effectiveness still requires broader matched comparisons (Appendix~\ref{app:prospective-gate}).

\subsection{Adapter readiness and treatment scope}
ToolSandbox's 300/300 constructor/evaluator audit checks construction without model calls. Adapter-level checks are reported separately from generated-update effectiveness.

TimeWarp WikiTest produced 24 Qwen/AgentLab run records with a 40-step cap: all 24 were truncated and none completed the task within budget. These are unsuccessful generic-agent baseline runs, not a CoordinateClosure comparison. Truncation alone does not distinguish inadequate budget from navigation, tool, or reasoning failures.
AgentDojo currently has 20 native-bound events and seven excluded events among 27 checks; SkillsBench's citation-check smoke verifies adapter readiness only.

\subsection{Public checks and repair for small workflows}
\label{app:workflow-repair}
We construct small executable workflows from four ToolSandbox function families: reminders, contacts, messaging, and network settings. Each family has a single-change and a coupled-change condition, yielding eight conditions involving arguments and return structures, time units, or resource prerequisites. These are synthetic controlled changes, not historical upstream migrations. Each condition is proposed by \texttt{qwen-plus} and \texttt{glm-5.2} (temperature zero, requested seed 11, 2,048 output-token limit), giving 16 condition--model units per format. All families and variants are development data. Admission and final replay use distinct generated task entities, not source-family-held-out data. Execution invokes native tool functions rather than a full conversational agent.

\begin{table}[!htb]
\centering
\small
\begin{tabular}{lrrr}
\toprule
Initial format & \shortstack{First-pass\\target} & \shortstack{After common\\checks/repair/gate} & Total tokens \\
\midrule
Whole rewrite & 16/16 & 16/16 & 11,008 \\
Untyped patch & 7/16 & 16/16 & 19,189 \\
Coordinate patch & 6/16 & 16/16 & 20,977 \\
\bottomrule
\end{tabular}
\caption{Small-workflow development diagnostic. All groups use the same public checker and at most one generated repair after public rejection, followed by admission and final replay. Tokens include reported input/output usage for initial generation and repair.}
\label{tab:workflow-repair}
\end{table}

The public checker verifies tool arguments, unit conversions, output references, resource prerequisites, and protected scope. It uses hand-declared rules and a fixed dependency structure specialized to these families, rather than discovering complete semantic dependencies for new tasks; exchanging two same-type inputs can still pass static checks. Repair receives only the public contract, candidate, and public-check errors, never final-task outcomes. With the common checker, at most one repair, and replay gate, all groups reach 16/16 on both target and protected tasks. Protection covers only an unchanged contact query, not broad capability retention. Ten repair requests were rejected before generation because of JSON-output configuration and each was retried once in a separately recorded recovery; original requests and gates remain preserved. The ledger contains 77 requests, 67 generated responses, and 51,174 reported tokens. Ten rejected requests have no usage records and are not treated as known zero-cost requests.

Common public checks and repair remove the final-success difference in these small workflows. Coordinate patches offer no success or token advantage here. The checker was refined during development; this diagnostic covers smaller artifacts and different cost components from the main maintenance study.

\subsection{Additional numerical-tool migrations}
Two previously unused source-scenario-derived numerical subtasks test timestamp differences and geographic distances. Synthetic API migrations change endpoint names, argument units or nesting, and return structures; upstream native functions perform the computation. Holiday and location retrieval are replaced by explicit public inputs. These are two events from one ToolSandbox utilities module, not complete conversations or independent source families. Contracts, scoring, and source hashes are frozen before generation. Twenty reference checks establish pre-migration validity, post-migration reference success, stale-target failure, and unchanged unit-conversion behavior.

Two backbones (\texttt{qwen-plus}, \texttt{glm-5.2}) generate three formats for both events: 12 candidates with temperature zero, requested seed 11, and a 2,048-token output limit. A fourth arm reuses the identical coordinate candidate with replay admission and no repair. Admission and final tasks use distinct numerical inputs. All eight admission and 32 final execution receipts are present, with no binding errors or unknown final outcomes. Whole rewriting, untyped patches, coordinate patches, and coordinate patches with admission pass respectively 3/4, 2/4, 3/4, and 3/4 target tasks; all four pass 4/4 protected unit-conversion tasks. Each denominator is two events times two models. Generated tokens total 2,270, 1,914, and 2,009 for the three proposal formats; admission reuses its candidate. An incorrect return-value selector causes all three GLM distance candidates to fail; rejection retains a stale endpoint and does not restore the target. Coordinate edits match rewriting in this small development comparison, with retention limited to the protected unit-conversion task.

\paragraph{Public-schema repair extension.}
A subsequent development version reuses the 12 candidates above and adds 12 fresh candidates for two date/time conversions from the same utilities module. All three formats receive the same public-schema checker and at most one repair; only the fourth arm additionally gates execution of the identical typed candidate. Forty reference checks pass before generation. The 16 admission slots comprise 12 executions and four public-check rejections; all 64 final slots are executed, yielding 46 passes and 18 failures. On the two reused events, target/protected passes are respectively 3/4 and 4/4 for rewriting, 2/4 and 3/4 for untyped edits, 3/4 and 4/4 for typed edits, and 3/4 and 4/4 with admission. On the two fresh events these counts are 2/4 and 2/4, 3/4 and 3/4, 2/4 and 2/4, and 2/4 and 4/4. New generation and repair use 16,635 reported tokens; reused generation adds 6,193 historical tokens, counted once. Some responses return a public-packet wrapper instead of the required artifact and remain failures under the frozen parser. Admission preserves the parent for invalid candidates, protecting the unchanged task without restoring the changed target. One reused untyped candidate loses executability after repair. This version was designed after inspecting earlier development results; the schema checker is not a complete semantic dependency compiler, and the fresh events are not independent source-family tests.

\subsection{Policy diagnostics}
\begin{figure}[!htb]
\centering
\includegraphics[width=\linewidth]{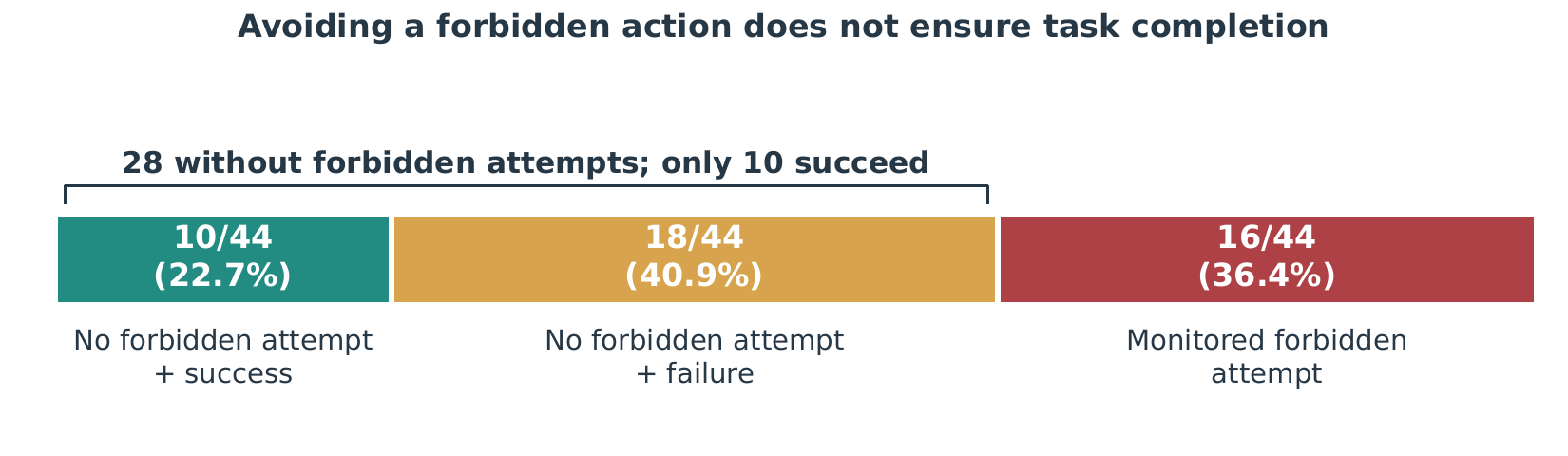}
\caption{Behavior and task completion in the frozen policy-state diagnostic. Among 44 observable updated changed-task trajectories, 28 contain no monitored forbidden attempt, but only 10 also complete the task. Four of 48 pairs are unobservable and excluded from this diagnostic breakdown. This is not a CoordinateClosure-versus-baseline comparison; the monitor covers specified forbidden actions only.}
\label{fig:compliance-completion}
\end{figure}
In the frozen $\tau^2$ v4b development cohort, 240 cells span four coupled events, 20 source tasks, two policy states, three trial seeds, and two backbones. Five cells remain operationally unknown. Of 48 changed-state pairs, 44 are observable on both sides. The paired effect on explicit violation attempts is $-0.545$, with 27 wins, three losses, and 14 ties; the source-cluster bootstrap interval is $[-0.813,-0.198]$. Changed-task native success has a paired effect of $+0.205$ with interval $[0,0.438]$, and protected retention has effect $-0.014$ with interval $[-0.097,0.069]$. This stale-versus-updated-policy comparison is development evidence, not a four-method update comparison.

An independent 32-cell accepted-policy diagnostic yields 31 completed trajectories and one operational unknown. The scorer omits required natural-language assertions and three changed events have no expected database actions. Thus component passes cannot establish full task recovery. Qwen3.8-Max and GLM-5.2 independently review 52 assertions in 25 packets, agreeing on 50 (31 passes and 19 failures), with two disputes. The 96.15\% agreement is not calibrated accuracy. An eight-trajectory exchange smoke remains affected by premature simulator termination: a database pass can coexist with no requested action. These runs are excluded from method-effectiveness claims.

\subsection{Failure analysis and adaptive verification}
\subsection*{Protected-task expansion}\label{app:retention-development}
We select seven original-payment return tasks whose source users do not overlap with the admission or final tasks used in the preceding return and exchange diagnostics. Original goals and reference actions remain unchanged. Selection follows source-task ID order after schema and reference-replay validation; one reference-replay failure is excluded and recorded before method execution. Two actors execute three generated-policy arms and a stale-policy control with one trial seed, yielding 56 planned cells and 55 saved trajectories. One GLM untyped-patch execution remains operationally unknown.
\begin{table}[!htb]
\centering\small
\caption{Protected return tasks: native database-check passes over seven planned tasks per cell. U denotes an operationally unknown result, retained in the denominator. These are component-level development results, not full-task success rates.}
\label{tab:retention-development}
\begin{tabular}{lcccc}
\toprule
Actor & Stale & Whole rewrite & Untyped patch & Typed patch \\
\midrule
Qwen-plus & 3/7 & 5/7 & 6/7 & 6/7 \\
GLM-5.2 & 7/7 & 7/7 & 6/7 + 1 U & 7/7 \\
\bottomrule
\end{tabular}
\end{table}
Typed-versus-stale pairs yield three wins, zero losses, and four ties for Qwen, and seven ties for GLM. Typed and untyped generators produce identical policy text on this event for both actors, so differences between their executions cannot identify a coordinate-representation effect. The small, single-event development sample supports a limited database-retention observation; broader event coverage and statistical non-inferiority remain unverified.

\paragraph{Post-hoc semantic audit.} Two judges assessed all 55 saved trajectories; one planned cell had no trajectory. Of 110 initial judgments, three had invalid structure and received one retry each, without retrying valid disagreements or unfavorable judgments. Combining the effective judgments with database and necessary-action checks yields 39 pass, 6 fail, and 11 unknown outcomes over 56 planned cells. Typed-versus-stale pairs give Qwen one win, zero losses, four ties, and two unknown pairs; GLM has zero wins, one loss, four ties, and two unknown pairs. In the GLM loss, the requested savings were reported as \$38.32 instead of \$41.64 despite a successful database action. Thus, database retention alone does not establish semantic retention. These model-assisted, post-hoc diagnostics do not establish representation effects or calibrated evaluation accuracy.

\paragraph{Airline shadow evaluation.} A separate fixed-format evaluation produced 46 structurally valid judgments for 23 saved trajectories from 24 planned cells, without rerunning agents or changing the original admission decisions. The 20 final-evaluation cells yield 7 pass, 9 fail, and 4 unknown joint outcomes; one missing admission trajectory remains unknown in the full ledger. Both original gates retain their parent policy. Structural validity measures judgment-format completeness, not correctness; judge disagreement and uncertain simulator fidelity remain unresolved. This single-event development diagnostic is not a held-out method comparison.

\paragraph{Additional protected tasks and environment correction.}
Three additional source tasks, disjoint from the seven above, yield 24 complete trajectories across four policy states and two actors: 12 pass, four fail, and eight remain unknown under joint checks. One task exposes an upstream multi-item tool defect, so these original-environment results are not clean method-effectiveness evidence. A versioned correction is applied uniformly to all eight actor--policy combinations for that task; seven produce complete trajectories, with four passes, two failures, and two unknowns over the eight planned runs. Required-action checks accept equivalent item orderings while preserving item pairing, payment parameters, and multiplicity. Original receipts and outcomes are retained. The two task sets cover ten protected tasks for one policy-change event; the repaired batch repeats one task and is reported separately, not pooled as additional independent evidence. Source-database verification also confirms the earlier GLM savings regression: the cheapest valid replacements save \$14.18 and \$27.46, totaling \$41.64, rather than the reported \$38.32.

\paragraph{Retention diagnostic on additional source tasks.}
For the same policy event, source-ID selection adds six previously unused protected tasks (20, 33, 36, 40, 41, 43), preserving complete user goals and reusing fixed policy candidates. Two actors and four policy states run in the version-corrected environment, yielding 48 planned cells. Joint database, necessary-action, and two-judge semantic checks yield 23 pass, 6 fail, and 19 unknown outcomes; disagreements and uncertain user-simulator fidelity remain unknown. This denominator pools stale, whole-rewrite, untyped-local, and coordinate-update states and is not any method's success rate. The additional tasks broaden protected-task coverage for one event without adding independent change events. This development diagnostic is reported separately from the preceding batches and does not establish non-degradation.

\paragraph{Operational recovery and scoring review.}
The original 48-cell result remains unchanged. Six GLM trajectories interrupted by empty-assistant-message validation were each retried once under a fixed selection rule, retaining tasks, candidates, seeds, and scoring protocol. All six completed, with two-judge outcomes of three passes, two failures, and one unknown. Combining them with the 42 unchanged outcomes gives a secondary analysis of 26 passes, eight failures, and 14 unknowns. Recovery execution records an additional 739,916 input/output tokens; interrupted original prefixes and subsequent judgments incur separate costs. This secondary view does not replace the original 23/6/19 record.

A uniform offline evidence audit attributes the 14 historical unknown labels to 12 judge disagreements and two assertion-coverage errors. Twelve of these trajectories already fail database or necessary-action checks. For the conjunctive endpoint, any confirmed required-component failure takes precedence over unresolved judgments; unknown applies only when no component is confirmed to fail and at least one remains unresolved. The existing component-dominates reaggregation therefore yields 26 pass, 20 fail, and two unknown outcomes on the same 48 recovery-sensitivity trajectories. The historical judge-first counts (26/8/14) and all original judgments remain archived; no trajectory is rerun and the success numerator is unchanged. Localized judge errors include misreading available products and missing simulator departures from the source goal; these diagnoses do not justify selective reassignment to pass.

\paragraph{Synthetic recoverable policy event.}
A separately frozen, self-authored refund event executes a legal original-payment refund using actual $\tau^2$ RetailTools and its database. It is counted separately from the 12 ToolSandbox development conditions and does not alter the original unsatisfiable gift-card task. Admission and final checks use distinct synthetic entities, with reference controls verified before generation. Two models (Qwen-plus and GLM-5.2), two repeats (requested seeds 11 and 29), and three generation formats produce 12 candidates; the fourth arm reuses the typed candidate with replay admission and no repair. All 12 outputs materialize to the same updated artifact. All eight admission and 32 final executions pass; each method--model group has target and protected scores of 2/2. Offline replay exactly matches all 40 execution receipts. Generation uses 800, 828, and 884 tokens for the three formats, totaling 2,512. This single-mechanism diagnostic verifies a learned-update-to-execution path, with identical artifacts and tied outcomes.

\subsection*{Task-disjoint end-to-end admission diagnostic}\label{app:prospective-gate}
We connect candidate generation, native execution, independent admission replay, and final evaluation on one retail policy event. Qwen-plus executes tasks; Qwen-plus and GLM-5.2 independently assess natural-language assertions and user-simulator fidelity. Admission source tasks (58, 94) are disjoint from final source tasks (45, 75). This is a single-trial development diagnostic. Task45 permits a legal settlement alternative when a gift card cannot be used, and its original reference action already uses PayPal. The stale policy can therefore succeed; this construction is not assumed to force a stale-policy failure.

Each method executes one affected and one protected task. Joint checks require native reward components, natural-language assertions, user fidelity, and required-action evidence. Action evidence joins an assistant tool call to a subsequent successful tool result by call ID and checks the action name and full arguments; promises cannot substitute for execution. This is a necessary condition, not a replacement for semantic or permission checks.

\begin{table}[!htb]
\centering\small
\caption{End-to-end development diagnostic on recoverable tasks. Each cell contains one trajectory; unresolved outcomes stay in the denominator and do not rank methods.}
\label{tab:native-development}
\begin{tabular}{lcc}
\toprule
Method & Affected task & Protected task \\
\midrule
Stale policy & Pass & Pass \\
Whole-policy rewrite & Pass & Pass \\
Untyped local patch & Pass & Unknown \\
Typed coordinate patch & Disputed & Unknown \\
Typed patch + replay admission & Pass (parent) & Pass (parent) \\
\bottomrule
\end{tabular}
\end{table}

All ten final trajectories contain required-action evidence: seven pass joint checks, one has an assertion disagreement, and two have incomplete scoring schemas. Both admission trajectories lack required execution and have incomplete judgments, so the gate retains the parent. Final passes in the gated arm therefore concern the parent. Typed and untyped candidates contain identical policy text; their dialogue differences cannot identify a representation effect.

Four synthetic known-label scorer controls were also assessed by both models. Of eight calls, seven returned schema-valid opinions and six correctly classified both task outcome and user fidelity. Qwen falsely reported success when the user stopped before execution; GLM returned an incomplete response on that case. This motivated the programmatic action check. These small controls do not estimate scoring accuracy on natural trajectories, and the diagnostics are excluded from the main effectiveness table.

\subsection*{Fixed-budget probe diagnostic}
\begin{figure}[!htb]
\centering
\includegraphics[width=\linewidth]{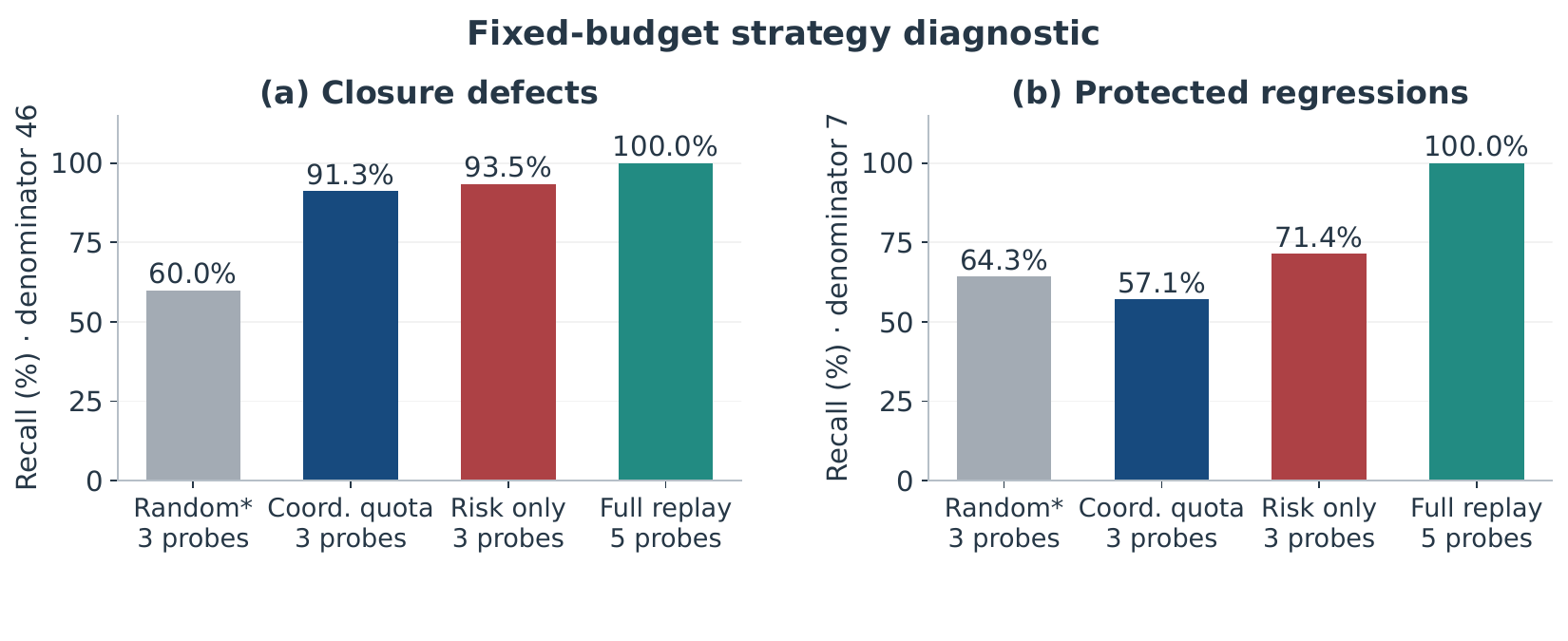}
\caption{Fixed-budget probe-selection diagnostics on 24 defective candidates. Three-probe strategies are compared with full five-probe replay. Defect recall uses 46 defects; protected-regression recall uses seven candidates. Random selection is an exact expectation. With no beneficial candidates, the pool cannot measure beneficial-update admission. This is not a budget sensitivity sweep.}
\label{fig:probe-budget}
\end{figure}
On a post-hoc pool of 24 defective candidates, each with two target and three protected probes, uniform selection of three probes gives 60.0\% expected defect recall. Risk-only selection achieves 93.5\%, compared with 91.3\% for a fixed two-target/one-protected coordinate quota. Protected-regression candidate recall is 71.4\% versus 57.1\%. This exposes a limitation of fixed quotas and motivates a risk--coverage selector, whose prospective benefit remains unverified. Since every candidate is defective, this pool cannot measure acceptance of beneficial updates.

In StreamingQA-derived case-disjoint development confirmation, whole/scoped annotation-state checks pass in 23/24 versus 24/24 cases for Qwen and 23/24 versus 23/24 for GLM; scoped updates use 26.6\% and 20.7\% fewer tokens in the respective reported comparisons. These checks are not native QA execution or verified factual evolution, and the small number of discordant outcomes does not establish an accuracy improvement. SkillFlow's 8/9 versus 8/9 outcome similarly provides a delivery-failure case, not evidence of improved recovery.

\section{Finite interface-dependency component ablation}
\label{app:mechanism-ablation}
The same four-arm ablation was executed on two existing ToolSandbox calculation fixtures, days until a holiday and geographic distance, with 17 candidates each. In each fixture, schema / dependency / replay / dependency-plus-replay yield target passes 2/2/1/1, protected passes 9/15/17/17, and joint passes 1/1/1/1, all out of 17. Jointly invalid candidates rejected number 0/12/15/15, with no rejection of the single beneficial candidate. Replay admission slots fall from 32 to 8; available-tool call attempts fall from 24 to 6, and successful calls from 20 to 6. Across both fixtures, 80 admission and 272 final workflow slots were actually executed with no model calls. These repeat constructed finite-contract tests on development fixtures, not independent natural events. Rejection retains the old artifact and therefore need not restore changed-task performance.

The checker verifies the public endpoint and two argument-name bindings. Each configuration retains the parent for a missing candidate. All 17 candidates remain in the denominator, and counts are compared within this versioned run.
\begin{table}[!htb]
\centering\small
\caption{Exhaustive candidates for one development interface event. Rejections are out of 15 jointly invalid candidates; target, protection and joint passes are out of 17. The sole valid candidate is never wrongly rejected.}
\begin{tabular}{lrrrrr}
\toprule
Checks & Reject & Target & Protect & Joint & Replays \\
\midrule
Schema & 0 & 2 & 9 & 1 & 0 \\
Schema + dependency & 12 & 2 & 15 & 1 & 0 \\
Schema + replay & 15 & 1 & 17 & 1 & 32 \\
Schema + dependency + replay & 15 & 1 & 17 & 1 & 8 \\
\bottomrule
\end{tabular}
\end{table}
Dependency filtering reduces candidate admission tool dispatches from 65 to 17; separately recorded reference dispatches decrease from 144 to 36. Final candidate dispatches remain 69 and final reference dispatches remain 85 in both replay configurations. The 75\% reduction concerns admission workflows, not total cost, latency, or tokens. Replay rejects one candidate that passes the target but fails protection, explaining the target-pass decrease from two to one. The injected suite does not estimate natural model-error frequencies; zero false rejections on one valid candidate is not a population estimate.

\section{Fresh change groups and dialogue-format diagnostics}
\label{app:latest-diagnostics}
\paragraph{New groups within the same source.}
The core release snapshot contains 19 development conditions; four additional IANA change groups are reported separately, not pooled into 23 independent source events. Two models and three generation formats produce 24 requests; the replay arm reuses coordinate candidates. Under a uniformly frozen metadata-compatibility rule, all 24 admission and 128 final conversions pass, with ties across methods. Strict-format results remain separate: 116 passes and 12 unknowns. A post-hoc semantic-diff audit finds exactly one affected zone-selector change in each of the 24 candidates and no unrelated edits. Generation costs vary by model and format, without a universal cost advantage.

\paragraph{Complete dialogues and observability revision.}
Local goal audits yield eight passes and four failures in the original 12-dialogue batch, versus seven passes, four failures and one unknown in the separate 12-dialogue error-budget sensitivity batch. These are not pooled into one success rate. The original latest-order task lacks an observable timestamp or ordering contract. We therefore construct a distinct revision with publicly queryable synthetic timestamps rather than rescore the old task. Both models complete authentication, timestamp and order lookup, and the correct payment answer before and after the policy change, yielding four goal-complete dialogues. This supports goal completion on one revised development task, not broad capability non-inferiority.

\paragraph{Execution-message normalization.}
All four revised-task traces contain raw messages mixing text with tool calls, contrary to the literal output protocol. The runner routes those messages to the tool environment without also delivering their text to the user. A new adapter clears only that text in the execution copy, preserving raw responses and model history, tool arguments and order; unsupported multiple calls are rejected rather than split. Offline native replay of all four traces, with 24 message projections, reproduces tool outputs, user-visible dialogue and final database states exactly. Normalized-protocol checks pass for all four traces and nine tests pass. Original literal failures remain reported. This is not a new online model run or a full-policy success rate.

\paragraph{Shared-artifact retail pilot.}
A separately frozen retail interface event uses admission tasks 60/62 and final tasks 79/90 from a different user group. Local-record exposure screening found no prior execution of the final group before this pilot; this is not a new upstream source. Two models generate three formats, with a fourth arm reusing the coordinate candidate for admission. Four provider errors among six initial requests are retained; a separately recorded connection-only recovery obtains the four missing responses without changing prompts or scoring. All six candidates materialize to the same artifact, and all 16 target/protected reference-workflow checks pass. Four complete Qwen protected dialogues subsequently complete the authorized cancellation; one misstates the price difference, so goal completion does not imply error-free communication. Four additional affected-task dialogues yield two original-goal passes and two failures. In the two failures, the simulated user invents a color inconsistent with the source task and database. These failures remain recorded and are flagged for simulator-fidelity failure, not attributed to updater differences. All four arms use identical artifacts. These evidence-indexed, non-blind assistant audits are diagnostic rather than calibrated ground truth; raw-message and normalized-transport compliance remain separate.

\paragraph{Airline interface pilot.}
One additional controlled baggage-interface event uses admission tasks 17/6 and final tasks 12/5 from two different final user groups, with local-record exposure checks and native reference/stale preflight tests. Three Qwen generation requests compare rewriting, untyped patches and coordinate patches; admission reuses the coordinate candidate. The latter two formats and the gated arm pass both final reference workflows, yielding six passes. Rewriting returns a normal model response but reverses the argument mapping and is rejected by the frozen materializer; its two execution outcomes remain unknown, not provider failures eligible for infrastructure recovery. This single-event format-conformance difference is not a reliable method ranking. Natural-language goal constraints are not evaluated by these reference workflows. Neither pilot increases the 19-condition core release count.

\paragraph{Repeated generation and contract sensitivity.}
On the already observed airline event, two models, four additional seeds, and three generation formats produce 24 requests; the gated arm reuses coordinate candidates. All eight generic-patch and eight coordinate-patch candidates materialize to the same artifact, whereas all eight whole-rewrite responses reverse the argument mapping. Of 64 final workflow slots, 48 pass and 16 remain unknown. These are correlated repetitions of one event, not additional independent events. A subsequent post-hoc sensitivity protocol, frozen before new calls, uniformly clarifies for all formats that mapping keys are old parameter names and values are new names. Six requests across two models, three formats, and another seed all materialize successfully; all 16 final four-arm slots pass, with identical candidates. Independent reconstruction without model calls matches candidates and final receipts in both batches. The clarified-contract result prevents interpreting the earlier difference as a coordinate-specific advantage. The batches use different seeds and are not a paired causal estimate of the prompt change. Generation/materialization errors must be distinguished from execution outcomes; this diagnostic shows no coordinate-patch token advantage.

\paragraph{Task-group-disjoint native workflow supplement.}
Four additional endpoint migrations use six final airline task/user groups, disjoint from five admission groups; local execution records contain no prior runs of these final groups. Twelve initial generations and eight public-feedback repairs consume 41,376 tokens. Under the frozen materialization contract, rewriting passes all four joint event checks; generic and coordinate patches each yield eight unknown final workflow outcomes, while the admission arm retains the parent, passing four protected workflows and failing four affected workflows. Both patch formats assign an entire argument-mapping object, whereas the materializer accepts only leaf assignments, a restriction omitted from the public format instructions. Independent reconstruction matches all original results. A separate post-hoc analysis expands object assignments into equivalent leaf assignments without changing any values or calling a model; all four arms then pass all four joint event checks (32/32 final workflow cells). The apparent ranking therefore depends on an underspecified representation contract. The fourth arm uses public binding checks and admission, not the full controller; reference-workflow replay does not evaluate complete conversations. All raw failures, unknowns and repair costs are retained.

\paragraph{Additional reproduction checks.}
Fresh extraction of the historical evidence archive verified all 1,727 payload hashes. An empty-cache IANA rebuild compiles the six core and four additional conditions. Existing-environment reruns reproduce 216 ToolSandbox controls and nine synthetic-retail outcomes, including two expected unknowns. These reproduce existing conditions; clean-environment results follow below.

\paragraph{Isolated native retail reconstruction.}
A further reconstruction downloads the fixed upstream commit and installs 74 version-pinned dependencies in a fresh Python 3.12 virtual environment without copying the previous environment. Seven native controls and two expected unknowns for the same synthetic policy event reproduce the original full receipt hashes. Runtime networking is disabled and no model is called. The entrypoint requires Python 3.12; verification covers local reconstruction of this existing policy fixture.

\paragraph{Isolated native ToolSandbox reconstruction.}
The existing 12-condition subset is also reconstructed from the fixed official source in a fresh Python 3.11 environment with 102 version-pinned dependencies. All 216 native controls match the frozen expectations: 120 reference/stale controls, 48 injected protection faults detected, and 48 corresponding unchanged-target passes, with no operational unknowns. A separate audit verifies all new receipt hashes and the complete historical control population. Execution networking is disabled and no model is called. These repeat existing controls and do not add events or method-comparison trajectories.

\end{document}